\documentclass[10pt,onecolumn,letterpaper]{article}
\usepackage{cvpr}
\usepackage[numbers,sort&compress]{natbib}
\usepackage{times}
\usepackage{eso-pic}

\usepackage{epsfig}
\usepackage{graphicx}
\usepackage{amsthm}
\usepackage{amsmath}
\usepackage{amssymb}
\usepackage[breaklinks=true,bookmarks=false]{hyperref}

\numberwithin{equation}{subsection}
\usepackage{datetime}

\usepackage[utf8]{inputenc} % allow utf-8 input
\usepackage[T1]{fontenc}    % use 8-bit T1 fonts
\usepackage{hyperref}       % hyperlinks
\usepackage{url}            % simple URL typesetting
\usepackage{booktabs}       % professional-quality tables
\usepackage{amsfonts}       % blackboard math symbols
\usepackage{nicefrac}       % compact symbols for 1/2, etc.
\usepackage{microtype}      % microtypography
\usepackage{xcolor}         % colors

\usepackage{graphicx}
\usepackage{amsmath} 
\usepackage{algorithm}
\usepackage{algorithmic}
\usepackage{colortbl} 
\usepackage{tcolorbox}

\usepackage{multirow}
\usepackage{array}
\usepackage{longtable}
\usepackage{adjustbox}
\usepackage{array}
\usepackage{lscape}
\usepackage{natbib}
\usepackage[a4paper,margin=1in,headheight=15pt,headsep=24pt]{geometry}

\usepackage{fancyhdr}

\cvprfinalcopy % *** Uncomment this line for the final submission
\def\cvprPaperID{****} % *** Enter the CVPR Paper ID here
\allowdisplaybreaks

\begin{document}
\lhead{}
\lfoot{\date{\today},\date{\currenttime}}
\rfoot{NGD for DL}

\title{Instruction Learning Paradigms: A Dual Perspective on White-box and Black-box LLMs}
\author{
% Yanwei Ren, Liu Liu, Baosheng Yu, Jiayan Qiu, Quan Chen
\textbf{Yanwei Ren}$^{1,2}$\quad
\textbf{Liu Liu}$^{1,2*}$\quad
\textbf{Baosheng Yu}$^{3}$\quad
\textbf{Jiayan Qiu}$^{4}$\quad
\textbf{Quan Chen}$^{5}$\\[0.5em]
$^1$School of Artificial Intelligence, Beihang University\\
$^2$Hangzhou International Innovation Institute, Beihang University\\
$^3$ Nanyang Technological University\quad
$^4$University of Leicester \\
$^5$Kuaishou Technology
}
\maketitle
\begingroup
\renewcommand\thefootnote{*}
\footnotetext{Corresponding author: \texttt{liuliubh@buaa.edu.cn}}
\endgroup
\begin{abstract}
Optimizing instructions for large language models (LLMs) is critical for harnessing their full potential in complex and diverse tasks. However, relying solely on white-box approaches demands extensive computational resources and offers limited representational capacity, while black-box models
% , though robust, 
can incur prohibitive financial costs. To address these challenges, we introduce a novel framework that seamlessly merges the strengths of both paradigms. Black-box models provide high-quality, diverse instruction initializations, and white-box models supply fine-grained interpretability through hidden states and output features. By enforcing a semantic similarity constraint, these components fuse into a unified high-dimensional representation that captures deep semantic and structural nuances, enabling an iterative optimization process to refine instruction quality and adaptability.
Extensive evaluations across a broad spectrum of tasks—ranging from complex reasoning to cross-lingual generalization—demonstrate that our approach consistently outperforms state-of-the-art baselines. This fusion of black-box initialization with advanced semantic refinement yields a scalable and efficient solution, paving the way for next-generation LLM-driven applications in diverse real-world scenarios.
The source code will be released soon.
\end{abstract}
\section{Introduction}
The application of large language models (LLMs) spans a diverse range of tasks and domains \cite{naveed2023comprehensive}, highlighting the increasing importance of developing effective and high-quality instructions \cite{zhang2024towards}. While modern LLMs exhibit robust capabilities in instruction-following, the performance of these models is highly sensitive to the quality of the instructions they receive. Poorly designed instructions can significantly limit their potential \cite{lou2024large}. Even for powerful closed-source models, such as the ChatGPT series, the quality of input instructions directly impacts the output quality \cite{white2024chatgpt}. This is particularly evident in domains such as medical question-answering systems, where clear, concise, and precise instructions are crucial for minimizing the risk of misdiagnosis \cite{hake2024quality}. Thus, optimizing instructions is essential for unlocking the full potential of LLMs and achieving reliable, high-quality outcomes. Additionally, with supervised fine-tuning (SFT) demonstrating considerable success in enhancing LLM performance, the construction of high-quality instruction datasets has become a focal point of contemporary research \cite{liu2024mftcoder}.

% \begin{figure}[t]
%     \centering
%     % \includegraphics[width=1\linewidth]{intinctAP.png}
%     \includegraphics[width=0.5\textwidth]{ExpFigure/IntroframeV4.png}
%     \caption{The framework of instruction optimization based on black-box and white-box model}
%     \label{fig:enter-label-sub}
% \end{figure}
% \setlength{\textfloatsep}{10pt}

Traditional approaches to instruction optimization typically involve the manual refinement of prompts to guide models toward more accurate outputs \cite{liu2023prompting}. However, this method is both cost-prohibitive and difficult to scale, which has driven a shift towards automated approaches. Despite advancements in automated prompt engineering, several fundamental challenges persist. Current solutions often treat prompt optimization as a black-box problem \cite{huang2024large}, relying on gradient-free search methods or heuristic-based techniques. These approaches are not only computationally expensive but also prone to converging at suboptimal local minima. Moreover, they frequently overlook the valuable intermediate representations within LLMs. Effectively leveraging these internal signals could offer more precise guidance throughout the optimization process. This raises a crucial question: how can we systematically integrate black-box feedback, such as evaluation scores, with white-box insights, such as hidden representations, to refine instructions and enhance performance?

The core insight driving our approach is that large language models (LLMs) encode critical information regarding instruction quality within their hidden layers \cite{liu2024mftcoder}. By integrating black-box feedback, such as scores from external evaluation models, with white-box access to the internal hidden representations, we establish a more intelligent and informed feedback loop for instruction learning. This combined approach allows us to more effectively guide the optimization process. Furthermore, an iterative framework ensures that each optimization step is informed by up-to-date performance signals, thereby minimizing dependence on random or naive search strategies.
The ultimate objective is to generate instructions that consistently produce higher-quality responses from black-box models while remaining adaptable across various application scenarios.

Therefore, we propose a novel instruction learning algorithm consists of three key stages 
% as shown in Figure \ref{fig:enter-label-sub}
. We firstly initialize the instruction dataset by harnessing the advanced capabilities of a cutting-edge black-box model, evaluating its performance and collecting hidden representations from a white-box model. Then. we train a custom neural network to predict instruction scores, encouraging the alignment of its representations with those of the highest-performing instructions. Finally, we iteratively refine the instructions by generating candidate prompts, extracting hidden features, and re-evaluating performance until convergence. This iterative process continues for a fixed number of iterations or until a predefined performance threshold is reached, ensuring steady convergence toward optimal instructions. The main contributions of our proposed framework are as follows:
\begin{itemize}
    \item 
    % \textbf{Contribution 1: A novel framework for instruction optimization.} 
    % \textbf{}
    We propose a novel framework for instruction learning that effectively enhances instruction generation performance across various tasks, including black-box and white-box model, i.e., ChatGPT and LLaMA.
    \item 
    % \textbf{Contribution 2: Three novel aspects of our framework.} 
    % First, our model outperforms existing methods like APE, InstructZero, EvoPrompt, and Instinct on key benchmarks (e.g., ``active\_to\_passive" and "cause\_and\_effect") with a notable improvement in the mean score across 30 tasks. Second, we introduce a refined mechanism for task-specific adaptation that boosts performance on diverse linguistic and reasoning tasks. Third, our framework incorporates a robust evaluation system, allowing for continuous refinement based on hidden representations and evaluation scores.
    We introduce a refined mechanism for task-specific adaptation that boosts performance on diverse linguistic and reasoning tasks. 
    % Besides, our framework 
    and incorporate a robust evaluation system, allowing for continuous refinement based on hidden representations and evaluation scores.
    \item 
    % \textbf{Contribution 3: Empirical validation.}
    Extensive experiments on a diverse set of tasks reveal that our framework consistently achieves the highest performance on several tasks, such as ``periodic\_elements"
    % , ``taxonomy\_animal", 
    and ``word\_sorting", as demonstrated in Table~\ref{Table-1}. The average score across 30 tasks demonstrates a clear improvement over existing methods, underscoring the effectiveness of our approach.
\end{itemize}

\section{Related works}

% \noindent
\textbf{Large Language Models and Instruction Tuning} 
With the continuous development and enhancement of computational resources, it has become possible to expand the number of parameters in pretrained models, enabling LLMs to achieve increasingly powerful capabilities \cite{zhao2021calibrate} . Taking the LLaMA model as an example, the number of parameters has increased from 65B (LLaMA 1) to 405B (LLaMA3.1) in just a few years \cite{zhao2023survey}. However, numerous studies have pointed out that simply increasing the number of model parameters no longer leads to significant performance improvements \cite{hoffmann2022training}. With the advent of instruction tuning \cite{zhao2021calibrate}, it has been found that instruction tuning can address some mismatch issues \cite{yu2023wavecoder}, allowing the model to better adapt to specific domains \cite{kung2023models} or tasks \cite{sun2023pushing}. Subsequent research has shifted the focus toward the construction of instruction datasets. Many studies have used their own custom-built data sets to fine-tune large models that exhibit strong performance, such as LIMA \cite{zhou2024lima}, Vicuna \cite{chiang2023vicuna}, and WizardLLM \cite{xu2023wizardlm}. These fine-tuned models have demonstrated exceptional capabilities.
% \\

% \noindent
\textbf{Instruction Construction}
The first method is manual construction, which involves data that is either manually annotated or directly sourced from the internet. Creating these datasets typically does not involve machine learning techniques and relies entirely on manual collection and validation, making them relatively small in size. For example, data sets such as Super-Natural Instructions \cite{wang2022super} and Alpaca \cite{taori2023alpaca} are high-quality data sets that have shown exceptional capabilities. However, the cost of manually constructing the data sets is very high and it is not feasible to scale them extensively.

With the rise of ChatGPT, researchers have attempted to use ChatGPT \cite{brown2020language} to replace humans in the dataset construction process. Based on ChatGPT, researchers have developed complex workflows to ensure the quality of the ChatGPT generated data sets \cite{longpre2023flan}. In addition to optimizing and constructing instruction data at the text level, some researchers have also explored the use of white-box large-language models to assist in instruction optimization and construction in high-dimensional spaces. For example, InstructZero \cite{chen2023instructzero}, compared to directly using ChatGPT for instruction optimization, appears more scalable in high-dimensional space exploration. This is especially important because cost is also a key consideration when building instruction datasets. Based on the InstructZero approach, INSTINCT further optimized the entire process using the UCB algorithm, reducing computational costs while achieving higher-quality instruction data \cite{linuse}.
\begin{figure*}[t]
    \centering
    \includegraphics[width=1.0\textwidth]{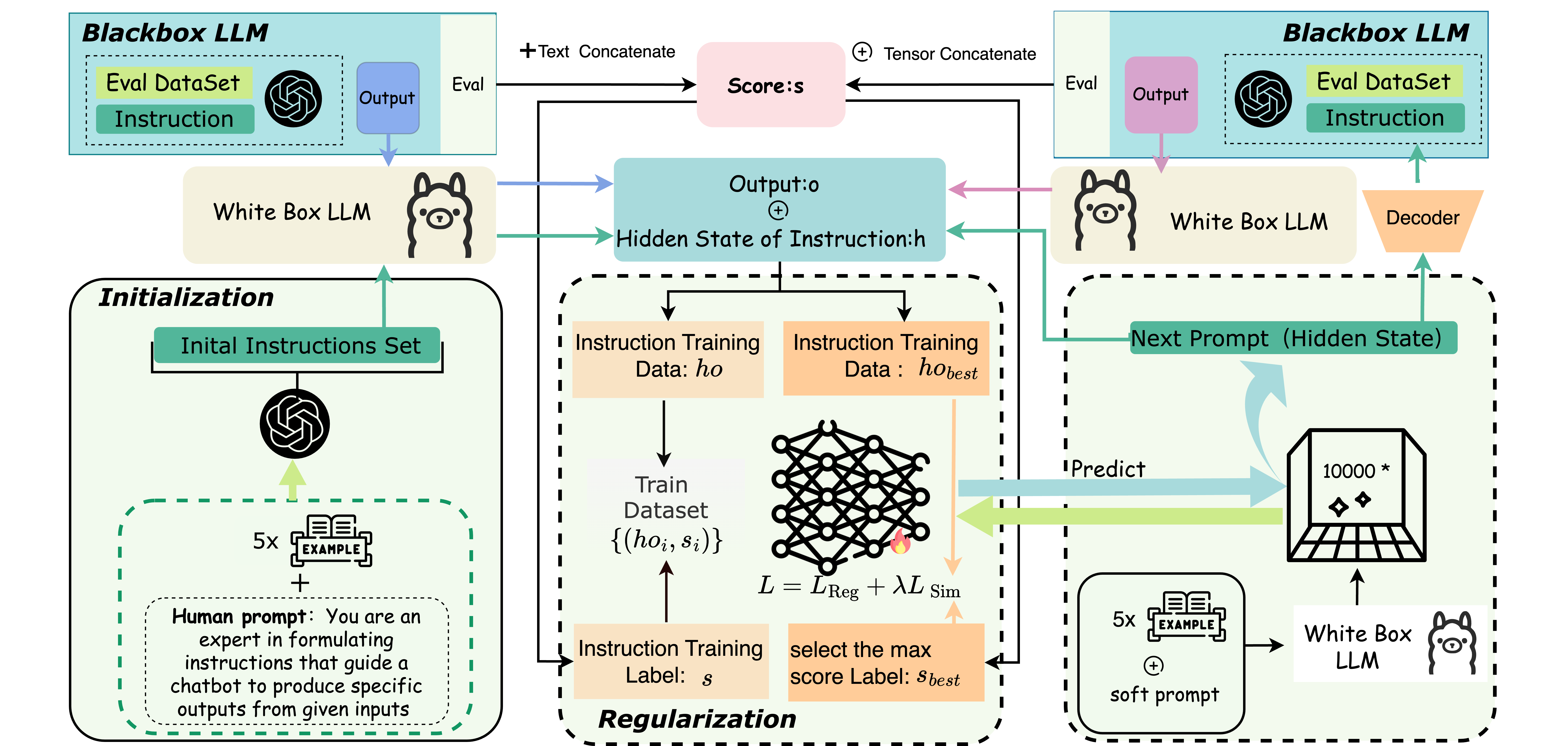}
    \caption{Overview of our instruction learning framework. The framework integrates black-box initialization for generating diverse, high-quality instruction candidates with white-box refinement, which leverages hidden representations and additional output representations for a more comprehensive optimization process. A similarity constraint aligns the instructions with ChatGPT outputs, ensuring semantic consistency, while iterative updates further enhance instruction effectiveness
    across diverse tasks.
    }
    \label{fig:enter-label}
\end{figure*}

\section{Methods}
In this section, we introduce a novel instruction framework designed to improve the instruction learning by leveraging both black-box model and white-box model as shown in Figure \ref{fig:enter-label}.
% The proposed framework  
% includes three components: 1) black-box initialization, 2) white-box instruction representation, and 3) Regularization of Instruction networks.

\subsection{Black-Box Initialization}

In previous methods \cite{zhou2022large,chen2023instructzero,linuse} for instruction learning, most approaches relied on embedding soft prompts and examples 
% (comprising pairs of inputs and outputs) 
into a white-box large language model to generate an initial set of instructions. However, this initialization process often led to several prominent issues. Specifically, the quality of the generated instructions heavily depended on the interpretative capabilities of the white-box model. 
Experimental results in Table~\ref{Table-1} show that while white-box models could readily produce effective instructions for simpler tasks, their performance deteriorated significantly with more complex tasks, e.g., second word letter.
This limitation can be attributed to the constrained parameter capacity and pretraining data of the white-box models, which often rendered them unable to fully comprehend the inputs during the initialization phase. As a result, ineffective initialization
% (e.g., generating null instructions) 
frequently occurred, leading to the failure of the entire instruction learning process.

With the continued development of black-box large language models (e.g., ChatGPT) and the decreasing costs of API usage, employing black-box models for initialization appears to be a more viable and effective solution. This approach not only addresses the limitations of white-box models but also mitigates the risk of catastrophic failure during the instruction learning process.
In this study, we utilize a carefully crafted prompt, i.e., ``\textit{You are an expert in formulating instructions that guide a chatbot to produce specific outputs from given inputs...}", along with a randomly selected set of five examples (comprising five input-output pairs) as input to ChatGPT. This approach enabled ChatGPT to generate an initial set of 40 instructions. By ensuring the quality of the initial instruction set while maintaining diversity among the instructions, we aim to provide the MLP with sufficient variability to learn the distinctions effectively. This variability allowed the model to better predict the scores associated with the corresponding soft prompts.

\subsection{White-box Instruction Representation}
To capture a more comprehensive set of instruction characteristics and strengthen the optimization process, we propose a simple yet effective fusion strategy. This strategy combines the hidden representations extracted from intermediate layers of a white-box model with its corresponding output representations, yielding a unified representation. By integrating the advantages of both feature types, this unified representation provides a richer and more stable foundation for further optimization and analysis, which incorporate three representations.
\\
% \noindent
\textbf{Hidden Representation ($h$)}
The hidden representation is extracted from the intermediate layers of the white-box model, capturing the deeper semantic structure and underlying information of the input instructions. Formally, it can be obtained as:
\begin{equation}
    h = \mathcal{M}_W(\pi), 
\end{equation}
where  $\mathcal{M}_W$ denotes  the function that extracts hidden features from the intermediate layers of the white-box model, and \(\pi\) denotes the input instruction.
\\
% \noindent
\textbf{Output Representation ($o$)}
The output representation directly reflects the generation result of the black-box model on the given input instruction, thereby encoding essential behavioral characteristics manifested during the execution process. It can be formulated as:
\begin{equation}
    o = \mathcal{M}_W(\mathcal{M}_{B}(\pi)_{\text{output}}),
\end{equation}
where $\mathcal{M}_B$ is the black-box model and $\mathcal{M}_{B}(\pi)_{\text{output}}$ means the final output of the balck-box model.
\\
\noindent
\textbf{Fusing the Representations}
To effectively harness the complementary information from both hidden and output representations, we first concatenate them into a single vector:
% \begin{equation}
    $ho = [h; o]$,
% \end{equation}
where \([h; o]\) denotes a dimensional concatenation of the two vectors. We then apply average pooling on the concatenated vector to reduce the feature dimensionality while smoothing out irrelevant fluctuations, resulting in a compact and robust fused representation:
\begin{equation}
    \widehat{ho} = \text{AvgPool}(ho),
\end{equation}
where $\text{AvgPool}(\cdot)$ is function of 
aggregating features from the hidden states of outpu and the instruction features by computing the mean over the specified dimensions.

\noindent
% \textbf{Advantages of the Fusion Strategy}
% This fusion strategy offers the following notable benefits:
% \begin{itemize}
%     \item \textit{Enhanced Representation Capability:} By integrating latent semantic information with generative outputs, the unified representation captures a richer set of instruction attributes.
%     \item \textit{Robustness:} The average pooling operation mitigates noise and reduces dimensionality, thus improving stability and generalization \cite{misiakiewicz2022learning}.

%     \item \textit{Efficiency:} Both concatenation and pooling incur low computational overhead, making this approach suitable for large-scale data scenarios.
% \end{itemize}
Empirical results demonstrate that the fused representation substantially boosts the model's prediction accuracy and lays a more stable foundation for subsequent iterative optimization processes.

\subsection{Regularization Instruction networks}
Given the varying sizes of the model parameters, we assume that the instructions generated by ChatGPT are effective for most tasks. Since the instructions generated by ChatGPT are in textual form, and cosine similarity can effectively capture semantic similarity between texts, we introduce cosine similarity in the instruction learning process. The key is to maintain semantic consistency between instructions. Using cosine similarity, we can ensure that the generated instructions align in semantic direction with the effective instructions generated by ChatGPT, thus improving the quality and effectiveness of the instructions during the optimization process. This helps ensure that the neural network (NN) can explore the next instruction based on the features of the effective instructions generated by ChatGPT.

The neural network $\mathcal{N}$ predicts the reward score $\hat{s}$ and n is the number of instruction for a given standardized contextual embedding ${ho}^*$:
% \begin{equation}
    $\hat{s} = \mathcal{N}({ho}^*)$. 
% \label{eq:nn_prediction}
% \end{equation}
%
% \noindent
% \textbf{Loss Functions}
The total loss $\mathcal{L}$ combines two components: the regression loss and the similarity loss:
\begin{equation}
    \mathcal{L} = \mathcal{L}_{\text{Reg}} + \lambda \mathcal{L}_{\text{Sim}},
\label{eq:total_loss}
\end{equation}
where $\lambda$ is the parameter, $\mathcal{L}_{\text{Reg}}$ and $\mathcal{L}_{\text{Sim}}$ are defined in Eq.\eqref{eq:regression_loss} and Eq.\eqref{eq:similarity_loss} as following
% \begin{itemize}
%     \item 

\noindent\textbf{Regression Loss}:
The regression loss minimizes the error between the predicted scores $\hat{s}_i$ and the true rewards $s_i^*$, $i\in[N+T]$ and $N$ is the number of initial instruction and $T$ is total iterations :
\begin{equation}
    \mathcal{L}_{\text{Reg}} = \frac{1}{N} \sum_{i=1}^N (\hat{s}_i - s_i^*)^2. \label{eq:regression_loss}
\end{equation}
% \item 
\noindent\textbf{Similarity regularization}:
To align contextual embeddings with the optimal instruction embedding, we calculate the cosine similarity between ${ho}_i^*$ and ${ho}_{\text{best}}^*$:
% \begin{equation}
$
    \tilde{s}_i = \frac{
    \langle{ho}_i^*,{ho}_{\text{best}}^*\rangle}{\|{ho}_i^*\| \|{ho}_{\text{best}}^*\|}
    $
    , 
% \label{eq:cosine_similarity}
% \end{equation}
where ${ho}_{\text{best}}^*$ is the hidden state of the best performance instruction
The similarity loss encourages consistent alignment:
\begin{equation}
\mathcal{L}_{\text{Sim}} = \frac{1}{N} \sum_{i=1}^N \big(\sigma(\hat{s}_i) - \sigma(\tilde{s}_i)\big)^2, \label{eq:similarity_loss}
\end{equation}
where $\sigma(\cdot)$ is the sigmoid function.
% \end{itemize}

In order to dynamically update the parameters $\lambda$, we introduce the \textit{Dynamic Weight Adjustment}. In particular, the weighting factor $\lambda$ is dynamically adjusted to balance the loss components:
% \begin{equation}
    $\lambda = \text{max}\left(a, \text{min}\left(b, {\mathcal{L}_{\text{Reg}}}\right)\right)$,
%     \label{eq:lambda}
% \end{equation}
where $a$ an $b$ are two parameters that control the weight of similarity loss.
% \begin{equation}
%     \lambda = \max\left(10^{-10}, \min\left(0.01, \frac{\mathcal{L}_{\text{Regression}}}{5}\right)\right). \label{eq:lambda}
% \end{equation}si
The total loss $\mathcal{L}$ is minimized using the Adam optimizer, with gradients computed via backpropagation. After iterative updates, the trained model parameters are saved for evaluation.

\noindent\textbf{Algorithm Description}
As shown in Algorithm \ref{alg:opt_algo}, Our framework consists of initialization and training. Black-box models generate a diverse instruction set, while white-box models extract hidden representations to train a neural network for predicting instruction effectiveness.
During training, the network iteratively refines soft prompts using black-box evaluation scores. A similarity constraint ensures semantic consistency, guiding optimization. The best instructions are continuously updated, balancing exploration and exploitation for improved instruction quality.

\begin{algorithm}[t]
\caption{Instruction Learning Algorithm under Black-box model and White-box model}
\label{alg:opt_algo}
\textbf{Input:} Dataset $\mathcal{D}$, Initial instruction dataset $\{\pi_i\}_{i=1}^N$ (e.g. ChatGPT, GPT4, GPT4o, OpenAI-o1), Black-box model $\mathcal{M}_B$ (for score evaluation), White-box model $\mathcal{M}_W$ (for hidden representations), Maximum iterations $T$\\
\textbf{Output:} Best instruction $\pi_{\text{best}}$

\begin{algorithmic}[1]
\STATE \textbf{Initialization}
\STATE Obtain evaluate scores $\{s_i\}_{i=1}^N$ through the initial instruction dataset and dataset based on $\mathcal{M}_B$.
\STATE Extract hidden representations $\{h_i, o_i\}_{i=1}^N$ using $\mathcal{M}_W$, and combine $h_i$ and $o_i$ to form $[h_i; o_i]$ and construct the instruction training dataset $\{([h_i; o_i], s_i)\}_{i=1}^N$.
% \STATE Initialize a three-layer neural network $NN$.
\STATE Train neural network $\mathcal{N}_s$ on the constructed instruction training dataset based on proposed loss function:
% \begin{itemize}
%     \item \textbf{RegressionLoss:} Measures the prediction error between $NN(ho_i)$ and $s_i$.
%     \item \textbf{SimilarityLoss:} Encourages alignment between $ho_i$ and the vector representation of the highest-scoring instruction.
% \end{itemize}
\STATE \textbf{Training}
\WHILE{$t \leq T$}
    \STATE Predict the next latent soft prompt $z_{t}$ using $\mathcal{N}_s$.
    \STATE Generate instruction $\pi_{t}$  using $\mathcal{M}_W$.
    % \STATE Extract hidden representations $h_{t}$ and $o_{t}$ using $\mathcal{M}_B$ .
    % \STATE Combine $h_{t}$ and $o_{t}$ to form $ho_{t}$.
    % \STATE Evaluate score $s_{t}$ using $BB(\pi_{t})$.
    % \STATE Update training dataset with $\{(ho_{t}, s_{t})\}$.
    \STATE Retrain $\mathcal{N}_s$ with the updated dataset.
    \STATE  Obtain evaluate scores $s_t$ through $\mathcal{M}_B$.
    \IF{$s_{t} > s_{\text{best}}$}
        \STATE Update $\pi_{\text{best}} \gets \pi_{t}$ and $s_{\text{best}} \gets s_{t}$.
    \ENDIF
\ENDWHILE
\STATE \textbf{return} $\pi_{\text{best}}$.
\end{algorithmic}
\end{algorithm}

\section{Experiments}

\subsection{Implementation}
We utilized ChatGPT-4o as the initialization model during the instruction learning process. To obtain the hidden states corresponding to the instructions and their outputs, we employed Vicuna-13B as a white-box model and applied average pooling to integrate the hidden states of the two components. For directional constraints, dynamically updated weights were introduced to impose constraints during the neural network training process. Additionally, we leveraged ChatGPT-3.5-Turbo to generate outputs corresponding to the given instructions. All experiments were conducted on a single NVIDIA L20 GPU.

\subsection{Datasets}
The 30 datasets used for instruction-guided tasks in our study are the same as those in InstructZero \cite{chen2023instructzero}. We excluded two datasets, CS Algorithms and ASCII, as the test datasets for these tasks have not been made publicly available. The entire dataset consists of 30 tasks, which cover a wide range of practical applications, including semantic analysis, sentiment judgment, code modification, and word localization tasks. This comprehensive coverage ensures a broad representation of real-world applications.

\subsection{Results}
\begin{table*}[t!]
\centering
\caption{Performance comparison over a benchmark of 30 tasks.}
\label{Table-1}
\renewcommand{\arraystretch}{1.0}
\begin{tabular}{lccccc}
\toprule
Tasks \& Algorithms 
& 
% \begin{tabular}[c]{@{}c@{}}APE\\ \cite{zhou2022large}\end{tabular} 
APE \cite{zhou2022large}
& 
% \begin{tabular}[c]{@{}c@{}}InstructZero\\ \cite{chen2023instructzero}\end{tabular} 
InstructZero \cite{chen2023instructzero}
& 
% \begin{tabular}[c]{@{}c@{}}EvoPrompt\\ \cite{guo2023connecting}\end{tabular} 
EvoPrompt \cite{guo2023connecting}
& 
% \begin{tabular}[c]{@{}c@{}}Instinct\\ \cite{linuse}\end{tabular} 
Instinct \cite{linuse}
& \textbf{ours} \\
\midrule
active\_to\_passive & \textbf{1.0000} & 0.9970 & 0.9930 & 0.9967 & \textbf{1.0000} \\
antonyms & 0.6370 & \textbf{0.8270} & 0.7970 & 0.7300 & 0.7533 \\
auto\_categorization & 0.2500 & \textbf{0.2570} & 0.2600 & 0.1900 & 0.2200 \\
auto\_debugging & 0.2920 & \textbf{0.3750} & \textbf{0.3750} & 0.2500 & 0.2917 \\
cause\_and\_effect & 0.5730 & 0.8130 & 0.8270 & 0.7467 & \textbf{0.9467} \\
common\_concept & 0.0690 & 0.0860 & \textbf{0.1210} & 0.0698 & 0.0956 \\
diff & 0.6730 & 0.6930 & \textbf{1.0000} & 0.5367 & 0.9300 \\
first\_word\_letter & \textbf{1.0000} & \textbf{1.0000} & \textbf{1.0000} & \textbf{1.0000} & \textbf{1.0000} \\
informal\_to\_formal & 0.5360 & 0.5310 & \textbf{0.6180} & 0.5098 & 0.5791 \\
larger\_animal & 0.8970 & \textbf{0.9000} & 0.7930 & \textbf{0.9033} & 0.8567 \\
letters\_list & \textbf{1.0000} & 0.5900 & \textbf{1.0000} & 0.9967 & 0.9933 \\
negation & 0.7530 & 0.7770 & 0.7870 & \textbf{0.7900} & 0.7633 \\
num\_to\_verbal & 0.9970 & \textbf{1.0000} & \textbf{1.0000} & \textbf{1.0000} & \textbf{1.0000} \\
object\_counting & \textbf{0.3630} & 0.3600 & 0.1170 & 0.3300 & 0.3433 \\
odd\_one\_out & 0.6330 & 0.6130 & 0.6530 & 0.3600 & \textbf{0.6800} \\
orthography\_starts\_with & 0.4570 & 0.5070 & \textbf{0.6000} & 0.5800 & 0.5167 \\
periodic\_elements & 0.9270 & 0.8670 & 0.9000 & \textbf{0.9400} & \textbf{0.9400} \\
rhymes & 0.1570 & \textbf{1.0000} & 0.6130 & 0.8567 & 0.6667 \\
second\_word\_letter & 0.7470 & 0.4330 & 0.4130 & 0.0667 & \textbf{0.9500} \\
sentence\_similarity & 0.0000 & \textbf{0.2830} & 0.1400 & 0.0000 & 0.0000 \\
sentiment & \textbf{0.9130} & 0.8770 & 0.8770 & 0.9067 & 0.8767 \\
singular\_to\_plural & \textbf{1.0000} & 0.9870 & \textbf{1.0000} & 0.9967 & 0.9900 \\
sum & 0.6730 & \textbf{1.0000} & \textbf{1.0000} & 0.9900 & 0.9800 \\
synonyms & \textbf{0.3600} & 0.2770 & 0.1370 & 0.1267 & 0.1567 \\
taxonomy\_animal & 0.3470 & \textbf{0.7170} & \textbf{0.7170} & 0.5300 & 0.6567 \\
translation\_en-de & \textbf{0.8400} & 0.8230 & 0.8130 & 0.6600 & 0.7900 \\
translation\_en-es & 0.8700 & \textbf{0.8730} & 0.8470 & \textbf{0.8733} & 0.8700 \\
translation\_en-fr & \textbf{0.8870} & 0.8770 & 0.8830 & 0.7233 & 0.8433 \\
word\_sorting & 0.3300 & 0.3100 & 0.5170 & 0.3267 & \textbf{0.5867} \\
word\_unscrambling & 0.4400 & 0.5500 & \textbf{0.6030} & 0.3533 & 0.5900 \\
\midrule
\rowcolor{gray!20} \textbf{MEAN} & 0.6207 & 0.6733 & 0.6800 & 0.6113 & \textbf{0.6955} \\
\bottomrule
\end{tabular}
\end{table*}
We evaluate our method against several state-of-the-art approaches, including APE \cite{zhou2022large}, InstructZero \cite{chen2023instructzero}, EvoPrompt \cite{guo2023connecting}, and Instinct \cite{linuse}, across a diverse set of 30 tasks. The results, summarized in Table~\ref{Table-1}, demonstrate that our method consistently outperforms all baselines, achieving the highest mean score of \textbf{0.6955}, compared to \textbf{0.6733} for InstructZero and \textbf{0.6800} for EvoPrompt.
\begin{figure*}[t!]

    \centering

    \includegraphics[width=0.32\textwidth]{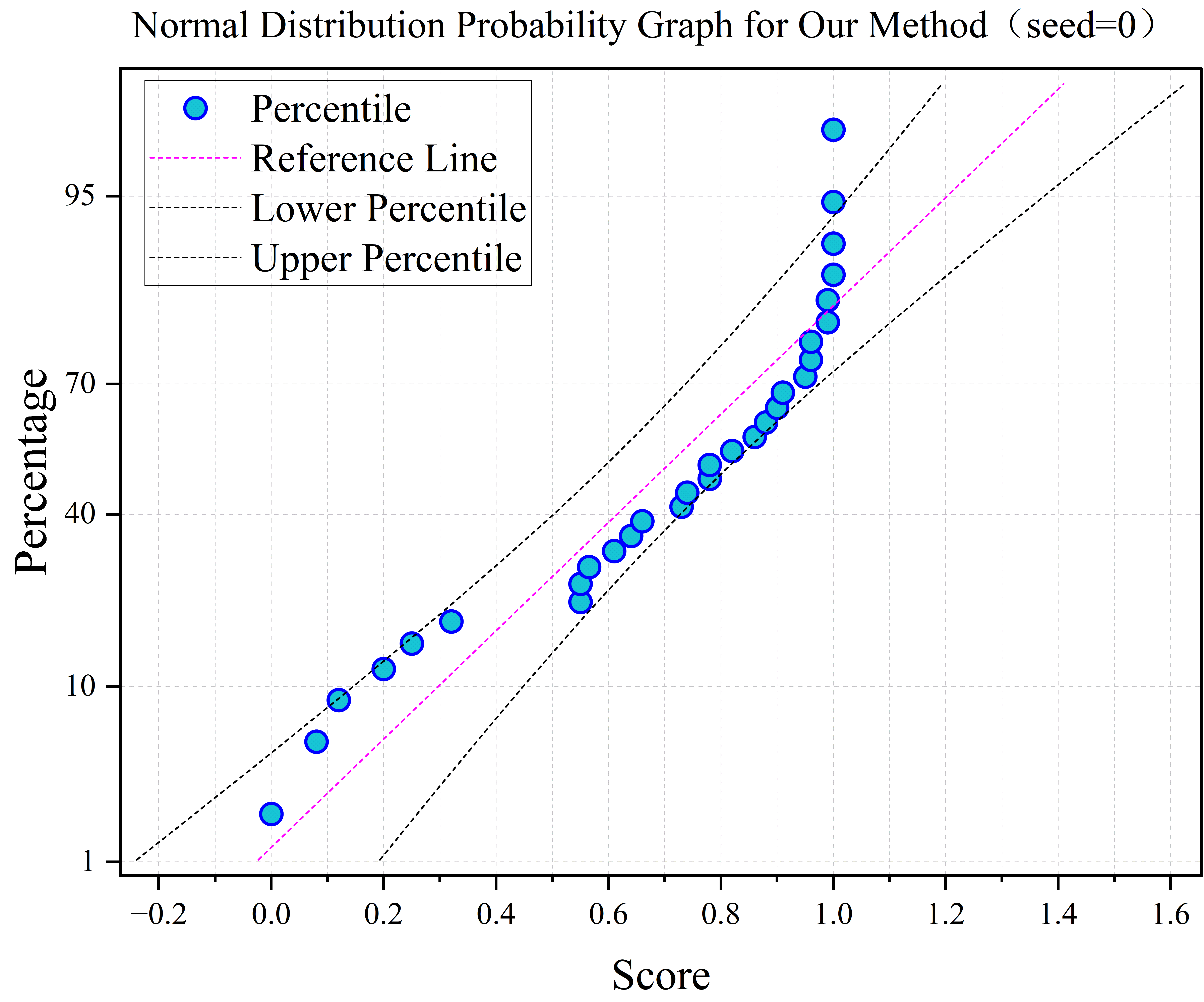}
    \includegraphics[width=0.32\textwidth]{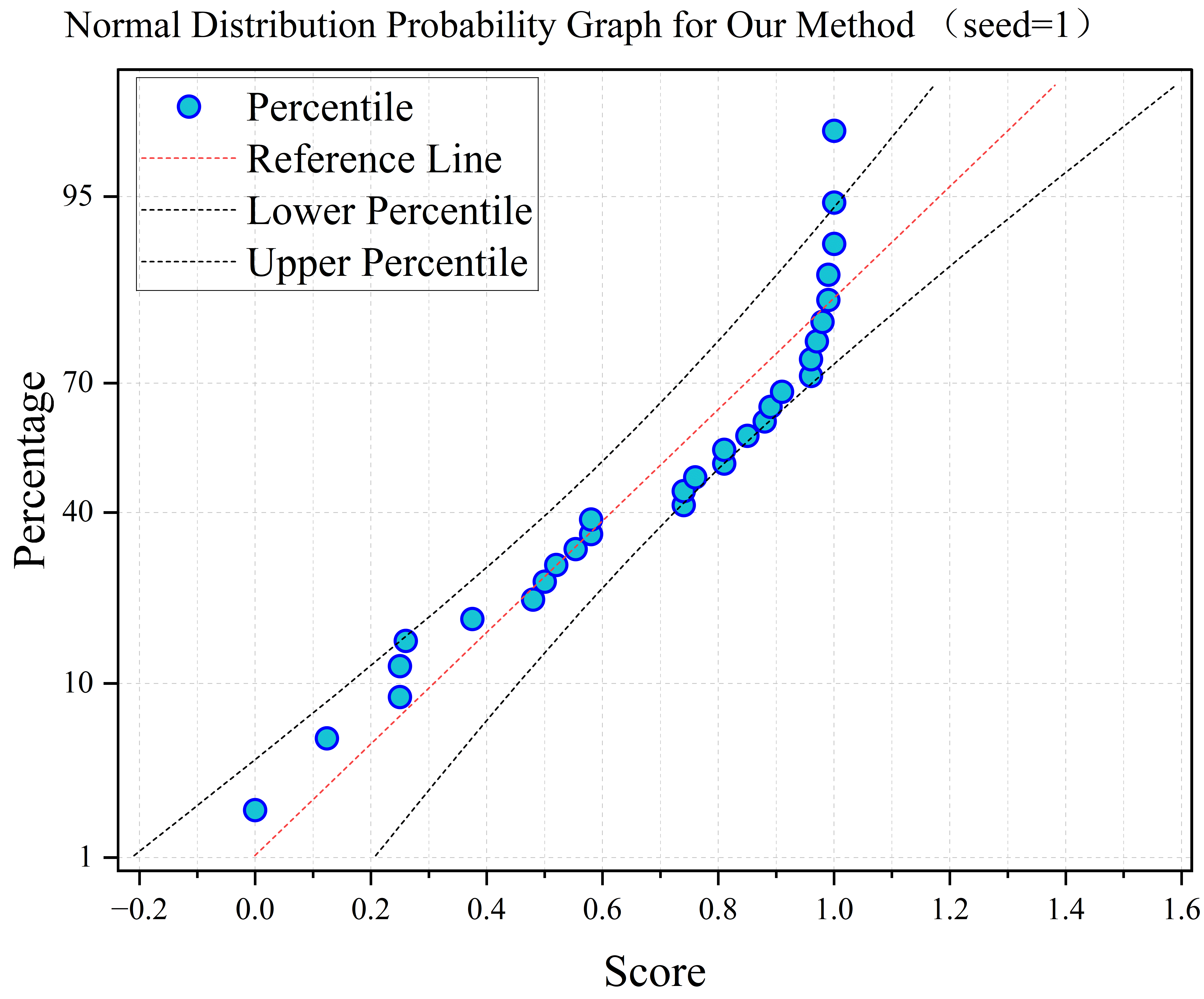}
    \includegraphics[width=0.32\textwidth]{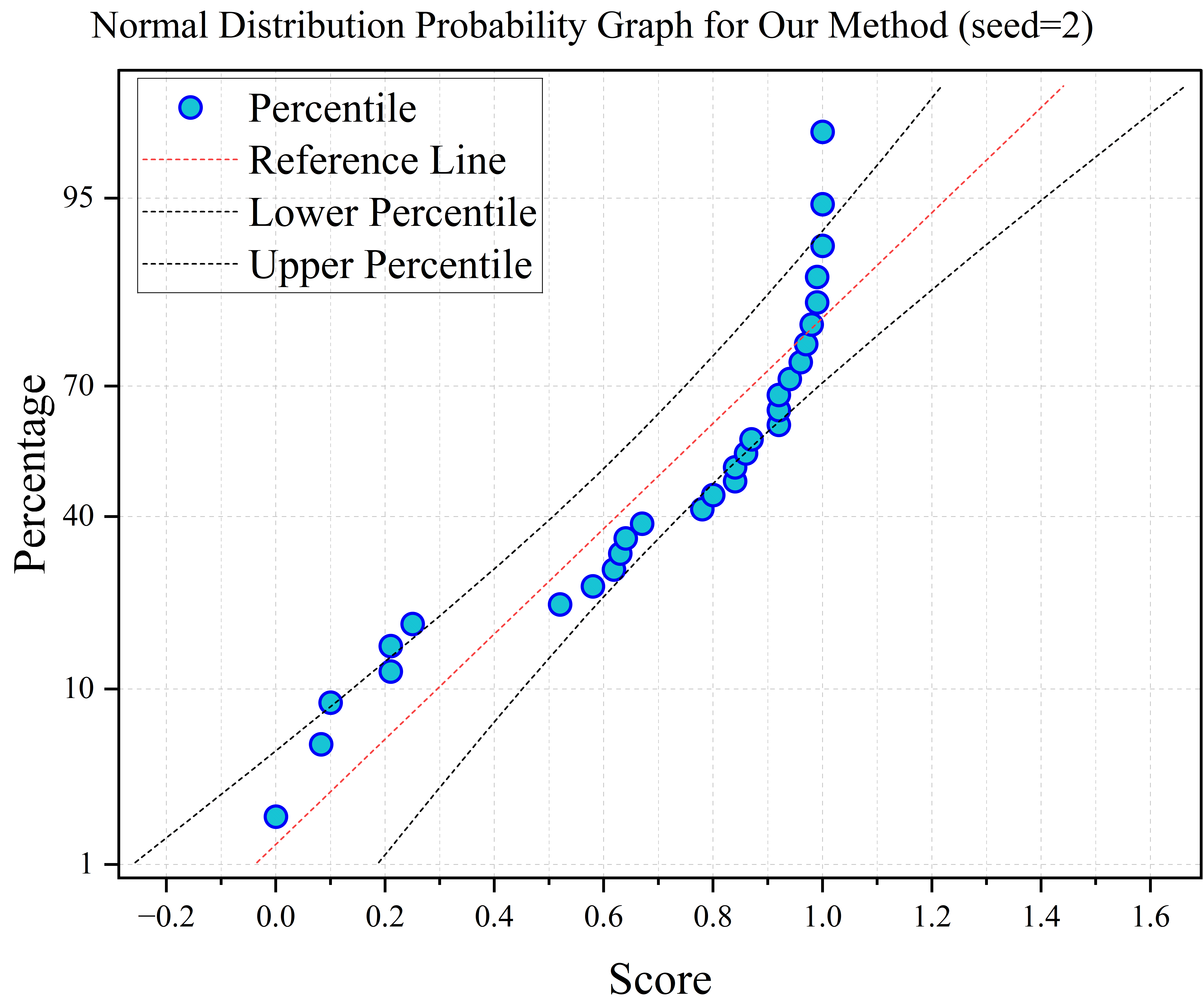} 
\caption{Normal Distribution Probability Graph for Our Method (3 seeds) }
\label{fig:3seedtest}
\end{figure*}
Figure~\ref{fig:3seedtest} shows the Normal Distribution Probability Graph for our method across three different seeds (0, 1, and 2). The plots illustrate the cumulative percentage of scores relative to the performance scores, with reference lines indicating the expected distribution. For all three seeds, the majority of the scores align closely with the reference line, suggesting that our method performs consistently and closely approximates the ideal distribution. The distribution of scores remains narrow, indicating stable performance with low variance across different initializations. This reinforces the robustness and reliability of our method.
Our approach excels in reasoning-intensive tasks, such as \textit{cause\_and\_effect}, where it achieves an impressive score of \textbf{0.9467}, significantly surpassing all other methods. Additionally, in simpler tasks like \textit{first\_word\_letter} (1.0000) and \textit{letters\_list} (1.0000), our method attains perfect scores, highlighting its strong performance across varying levels of task difficulty.

\begin{figure}[t]
    \centering
    \includegraphics[width=0.48\textwidth]{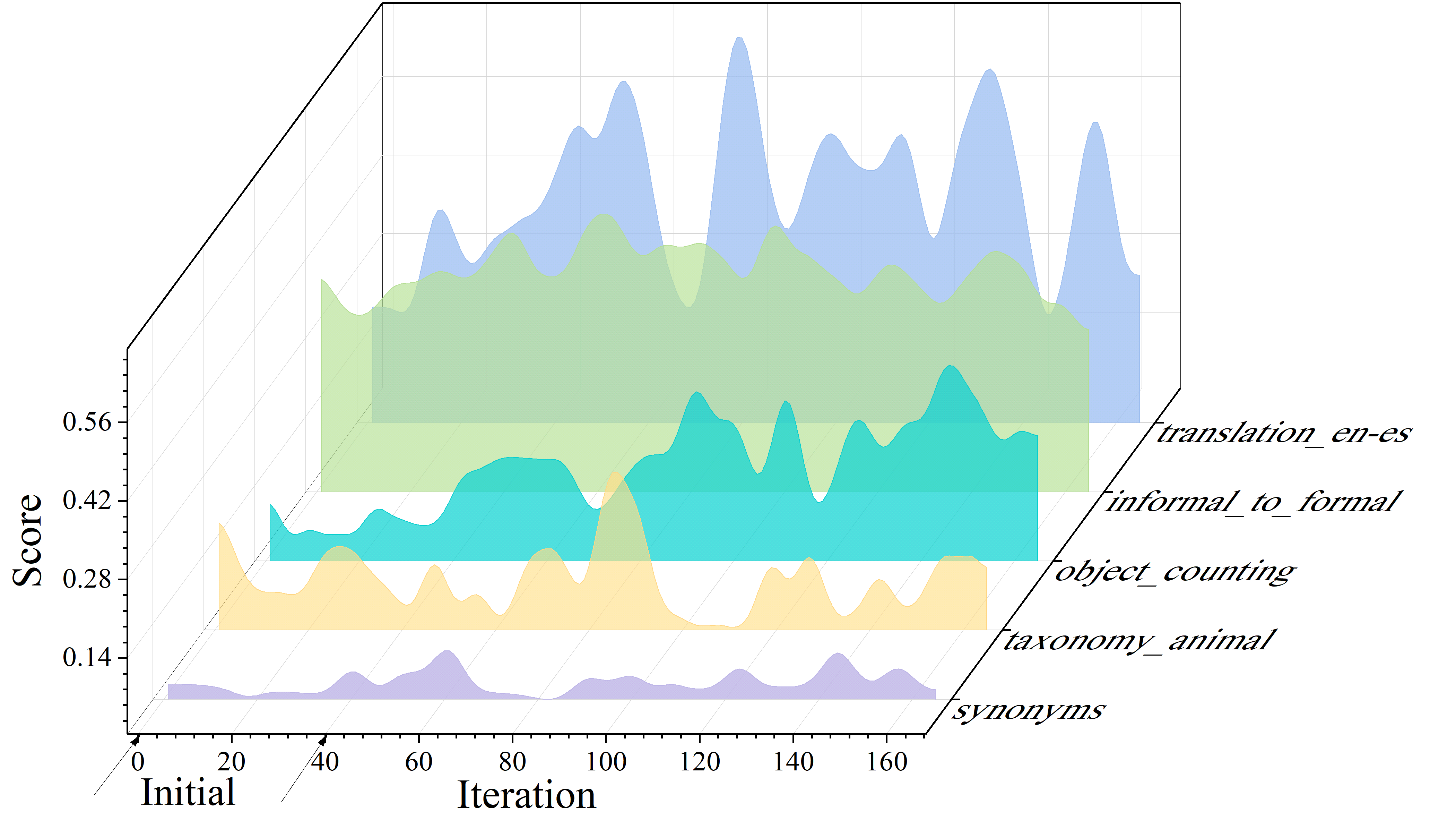}
    \includegraphics[width=0.48\textwidth]{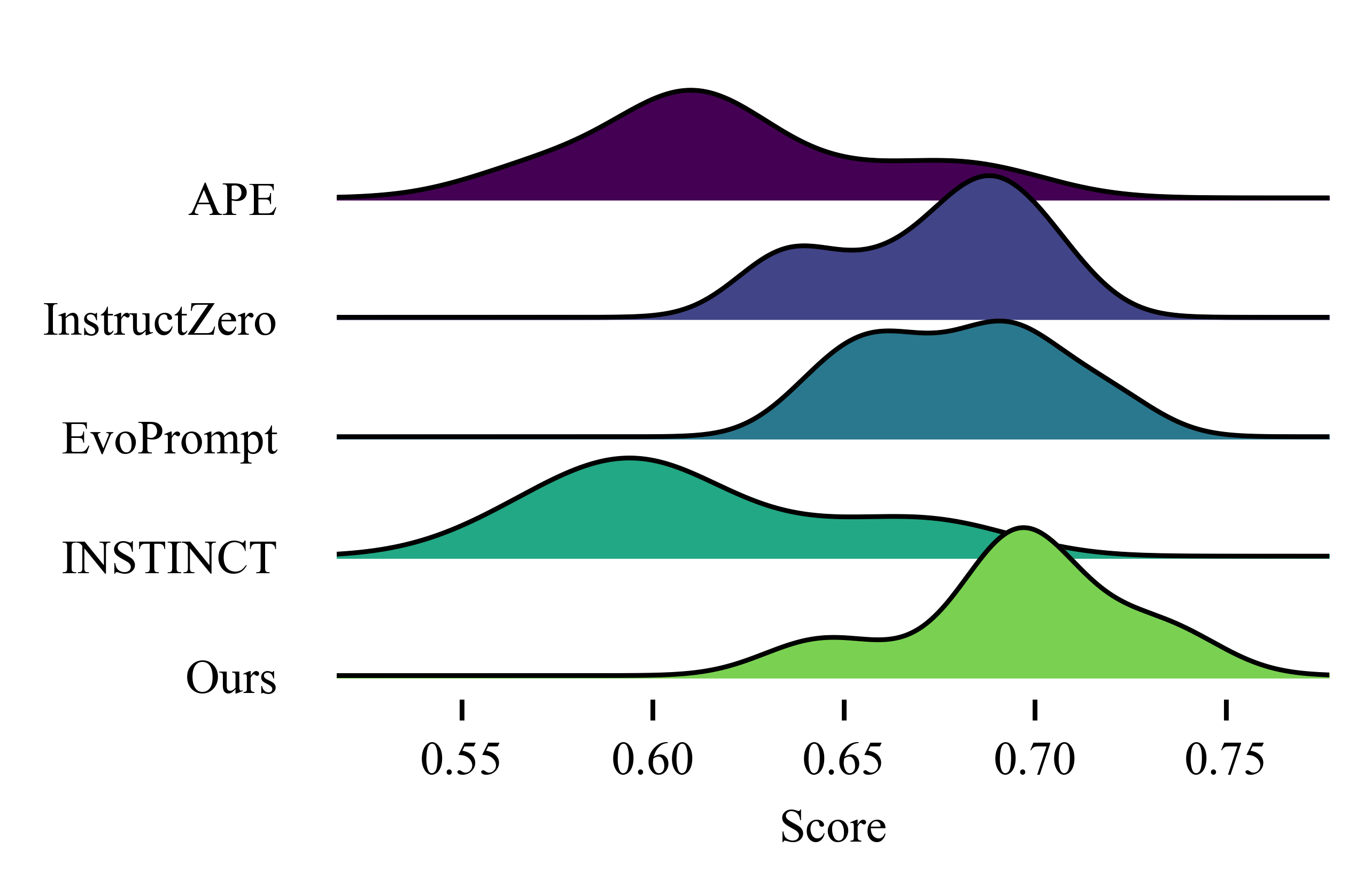}
    \vspace{0.3cm} 
    \caption{(Left)Evolution of instruction performance across iterations for 5 tasks, and (Right) Ridgeline plot showing the score distributions of different  methods across 30 tasks.
}
    \label{fig:3dwaterfall+jixiantu}
\end{figure}

To further assess the robustness of our approach, 
% Figure~\ref{fig:jixiantu} 
Figure~\ref{fig:3dwaterfall+jixiantu}
presents a ridgeline plot that illustrates the distribution of performance across tasks. Our method (denoted in green) consistently exhibits a higher density of high-performance scores, indicating not only superior performance but also greater consistency across a wide range of tasks.
%
% Figure~\ref{fig:3dwaterfall} 
Figure~\ref{fig:3dwaterfall+jixiantu}
depicts the evolution of instruction performance over iterations for selected tasks. While initial scores, generated by ChatGPT4o, are relatively low, the iterative optimization process significantly enhances performance, with our method surpassing the black-box LLM in subsequent iterations. This underscores the efficacy of our iterative refinement process in improving performance over time.

Finally,  a bubble chart  compares task performance across different baselines, highlighting performance across tasks of varying difficulty levels (More detailed information on task difficulty and the classification of tasks can be found in Table~\ref{table:task_difficulty} and Figure~\ref{fig:4bubbles} in Appendix). Our method performs exceptionally well across both simple and challenging tasks, with bubble size representing task difficulty. Notably, our method excels in the more difficult tasks, while maintaining stable performance in simpler ones, further validating its robustness across a spectrum of task complexities.
These results underscore the versatility and reliability of our approach, demonstrating its ability to handle both simple and complex tasks. By leveraging black-box initialization and iterative refinement, our method achieves superior performance across diverse NLP tasks, affirming its effectiveness for multi-task instruction generation.

\begin{table*}[t]
\centering
\caption{Performance comparison across 10 representative tasks in ablation experiments. The experiments evaluate the impact of different components and initialization settings, with ChatGPT 4o used as the initialization for our method.}
\label{tab:abaltionmaitab}
\setlength{\tabcolsep}{3pt} % 调整列间距
\begin{tabular}{@{}>{\raggedright}p{3cm}cccccc|c@{}}
\toprule
\multirow{3}{*}{\raisebox{-1.5ex}{{Tasks}}} & \multicolumn{4}{c}{Black-box LLM} & \multicolumn{1}{c}{Rep.} & \multicolumn{1}{c|}{Loss} & \multirow{3}{*}{\raisebox{-1.5ex}{Ours}} \\
\cmidrule(lr){2-5} \cmidrule(lr){6-6} \cmidrule(lr){7-7}
                       & ChatGPT & ChatGPT & ChatGPT & DeepSeek & Fuse Output  & Similarity &  \\
                       & (3.5)     & (4)       & (4o)      & (V3)       & Rep                 & Constraint &  \\
\midrule
active\_to\_passive    & 0.9833    & 0.9967    & 0.9633    & 1.0000 & 1.0000  & 0.9667    & 1.0000 \\
antonyms               & 0.7667    & 0.7633    & 0.7533    & 0.7533    & 0.8000  & 0.7567    & 0.7533 \\
cause\_and\_effect     & 0.6400    & 0.8800    & 0.8000    & 0.9200    & 0.9200    & 0.8133    & 0.9467 \\
diff                   & 0.6300    & 0.9967    & 0.9833    & 0.9600    & 0.9667    & 0.9567    & 0.9300 \\
informal\_to\_formal   & 0.4331    & 0.4733    & 0.4714    & 0.4709    & 0.5720  & 0.4816    & 0.5791 \\
object\_counting       & 0.3700    & 0.2167    & 0.3700    & 0.4367 & 0.2600    & 0.3100    & 0.3433 \\
orthography\_starts & 0.1200    & 0.5033    & 0.4567    & 0.1333    & 0.5267    & 0.6333    & 0.5167 \\
sentiment              & 0.8767    & 0.8967    & 0.8600    & 0.8933    & 0.8867    & 0.9100    & 0.8767 \\
translation\_en-de     & 0.5433    & 0.8467    & 0.8267    & 0.0433    & 0.8033    & 0.8033    & 0.7900 \\
word\_unscrambling     & 0.4533    & 0.4133    & 0.5533    & 0.5467    & 0.5733  & 0.5133    & 0.5900 \\
\midrule
\rowcolor[gray]{0.9}
\textbf{MEAN(30 Tasks)}          & \textbf{0.5645}    & \textbf{0.6698}    & \textbf{0.6713}    & \textbf{0.6299}    & \textbf{0.6717}    & \textbf{0.6840}    & \textbf{0.6955} \\
\bottomrule
\end{tabular}
\end{table*}

Overall, while our method exhibits some room for improvement in specific tasks such as stylistic transformation, it outperforms existing methods across a wide range of tasks. The superior performance in reasoning, syntactic transformation, and cross-lingual translation highlights the efficacy of our proposed method for multi-task instruction generation. These results demonstrate the potential of our approach to serve as a robust solution for diverse natural language processing tasks, providing a strong foundation for future improvements and extensions.

\subsection{Ablation experiment}
In order to assess the impact of key components in our instruction learning framework, we conduct a series of ablation experiments. These experiments systematically evaluate the contributions of individual factors, including initialization strategies, similarity regularization, output representation. By selectively removing or modifying these components, we gain insights into their respective roles in shaping the overall performance of the model across a diverse set of tasks (Table~\ref{tab:abaltionmaitab} presents the main results, with additional details provided in Table~\ref{tab:moredetails app} in Appendix.).

\begin{figure*}[t!]

    \centering
    \includegraphics[width=0.32\textwidth]{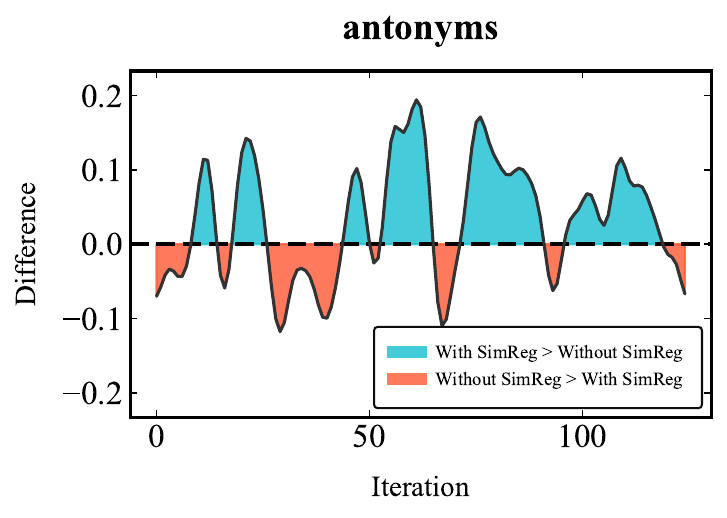}
    \includegraphics[width=0.32\textwidth]{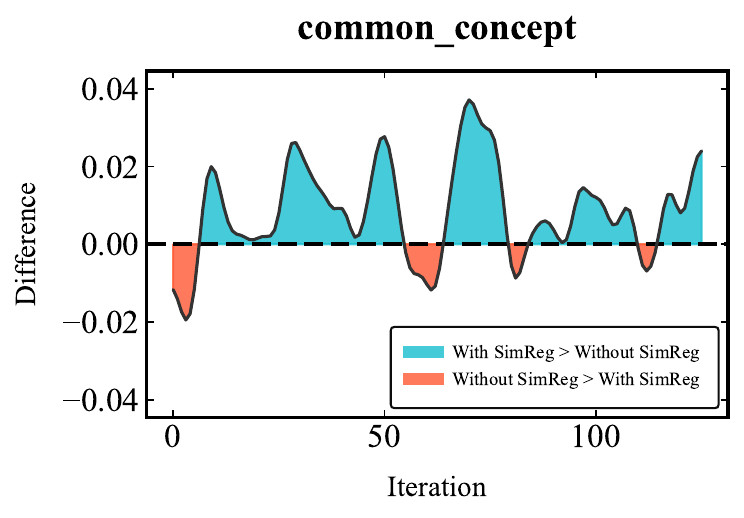}
    \includegraphics[width=0.32\textwidth]{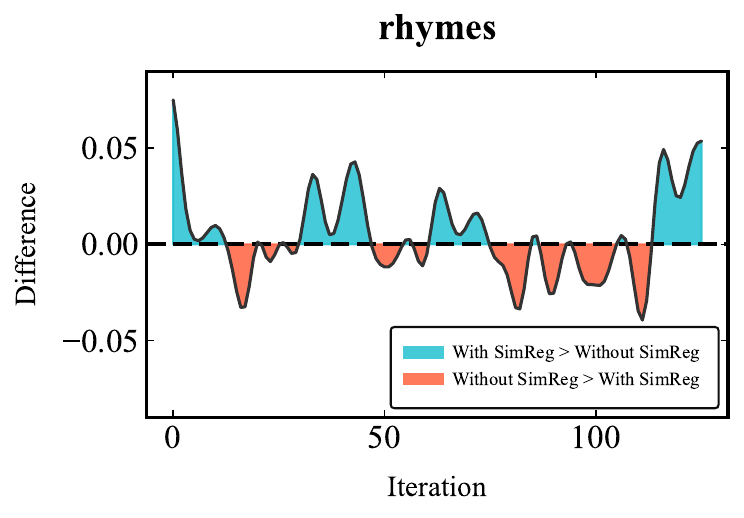} \\

    \includegraphics[width=0.32\textwidth]{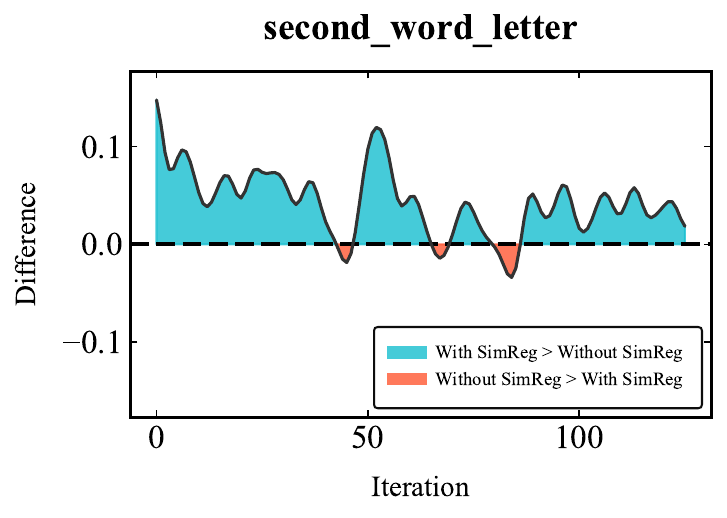}
    \includegraphics[width=0.32\textwidth]{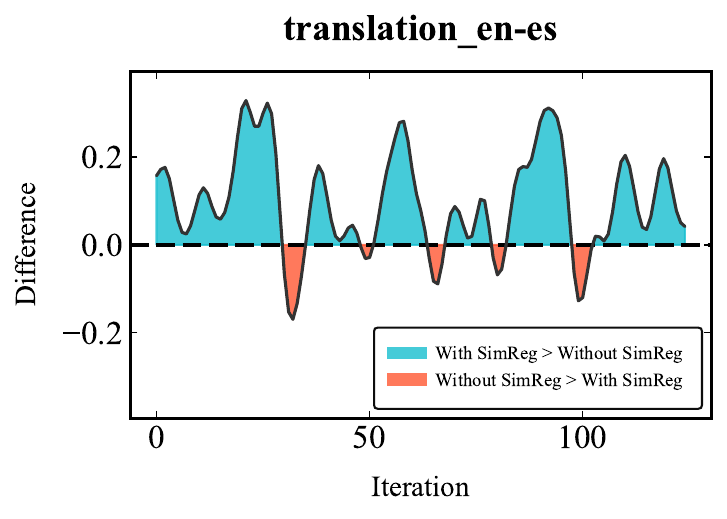}
    \includegraphics[width=0.32\textwidth]{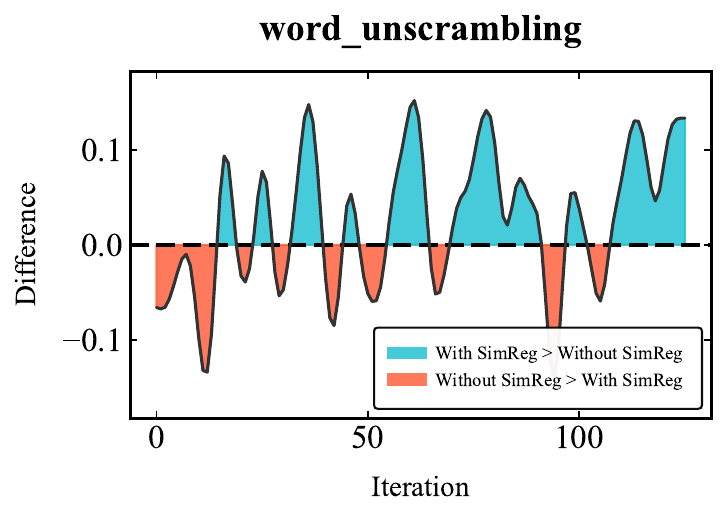}
    % \vspace{-0.3cm} 
    \caption{Performance difference between models with and without similarity regularization (SimReg) across 6 tasks. }
    \label{fig:ablation-loss1}
\end{figure*}

% \subsubsection

\textbf{The Validity of ChatGPT-Based Initialization}
We conduct an ablation study to evaluate the impact of black-box LLM initialization on instruction learning, comparing ChatGPT series (including version 3.5, 4, 4o) \cite{openai2023chatgpt} and DeepSeek V3 \cite{liu2024deepseek}. The results in Table~\ref{tab:abaltionmaitab} show that ChatGPT 4o achieves the highest mean score of 0.6713, surpassing ChatGPT 3.5 (0.5645) and ChatGPT 4 (0.6698). DeepSeek V3 performs well in certain tasks but has a lower overall mean (0.6299), suggesting that while different black-box models provide reasonable initializations, they lack consistency across tasks.

To further analyze the effectiveness of our approach, we integrate ChatGPT 4o initialization with white-box refinement and similarity constraints. Our method achieves a mean score of 0.6955, demonstrating a clear improvement over using ChatGPT alone. Notably, performance gains are most evident in structured reasoning tasks such as cause and effect (0.9467 vs. 0.8000 for ChatGPT 4o) and word unscrambling (0.5900 vs. 0.5533). These results confirm that while ChatGPT 4o provides strong initializations, our iterative refinement strategy enhances instruction quality beyond what black-box models alone can achieve.
These findings highlight the complementary role of black-box initialization and white-box optimization. By incorporating structural constraints and leveraging hidden representations, our approach systematically improves instruction effectiveness across diverse NLP tasks, reinforcing the necessity of a hybrid optimization framework over standalone black-box models.

% \subsubsection

% \noindent
\textbf{Similarity regularization
}
To evaluate the role of similarity regularization (SimReg) in instruction learning, we conduct an ablation study by removing the similarity constraint. The results indicate that the absence of SimReg leads to a noticeable degradation in instruction effectiveness. Tasks such as \textit{cause\_and\_effect} (0.9467 $ \to $ 0.8133) and \textit{informal\_to\_formal} (0.5791 $ \to $ 0.4816) exhibit significant drops, indicating that SimReg improves instruction alignment and enhances task adaptability.
The mean performance decreases from 0.6955 to 0.6840, suggesting that SimReg contributes to better generalization by reinforcing consistency in learned representations. This aligns with the results in 
% Appendix,
Figure~\ref{fig:ablation-loss1},
where models with SimReg exhibit a more stable performance trajectory across iterations, and Figure~\ref{fig:abaltion-loos2} in the Appendix , which shows that similarity constraints help reduce variability in instruction scores. These findings demonstrate that SimReg is particularly beneficial in tasks requiring fine-grained distinctions, as it ensures the model retains high-quality instruction structures 
via learning.
% throughout optimization.

% \subsubsection
% \noindent
\textbf{Ablation Study on Output Representation}
We investigate the impact of additional output representation by comparing models trained with and without it. Figure~\ref{fig:ablation--outrep} in Appendix shows the performance difference over iterations, where models with additional output representation consistently outperform those without. This improvement is particularly evident in tasks involving complex patterns such as \textit{letters\_list}, \textit{auto\_categorization}, and \textit{odd\_one\_out}. 
% as shown in Figure~\ref{fig:ablation--outrep}.
The green areas in the figures indicate where the model with output representation outperforms the model without it, confirming the stabilizing effect of the added representation.
Furthermore, Table~\ref{tab:abaltionmaitab} further quantifies the impact of additional output representation. Removing output representation results in performance drops in tasks like \textit{object\_counting} (0.3433 to 0.2600) and \textit{orthography\_starts\_with} (0.5267 to 0.5167), leading to a reduction in the overall mean score from 0.6955 to 0.6717. These results demonstrate that incorporating additional output representation helps the model leverage richer features, ensuring more stable and effective instruction optimization.
% \begin{figure}[t]
%     \centering
%     % \includegraphics[width=1\linewidth]{intinctAP.png}
%     \includegraphics[width=0.5\textwidth]{ExpFigure/jixiantu.png}
%     \vspace{-1cm} 
%     \caption{Ridgeline plot showing the score distributions of different  methods across 30 tasks.
% }

%     \label{fig:jixiantu}
% \end{figure}

\section{Conclusion}
In this work, we propose a novel instruction learning framework that merges the strengths of black-box models (high-quality, diverse initializations) and white-box models (fine-grained interpretability). By enforcing a semantic similarity constraint, our method unifies hidden state representations to iteratively refine instruction quality.
Extensive experiments across 30 tasks demonstrate consistent outperformance over state-of-the-art baselines, with an average performance of 0.6955. 
Looking ahead, we plan to reduce black-box dependencies and extend the framework to broader instruction-generation scenarios, including reinforcement learning and automated prompt engineering. By bridging black-box robustness and white-box interpretability, this work offers a promising direction for next-generation LLM-driven applications.

% {\bf Limitations} Our framework relies on a black-box model for initialization, which inherently introduces opacity and limits interpretability. Furthermore, we did not extend our experiments to larger and more diverse datasets, leaving the generalization of our approach to different tasks as an open question. Due to computational constraints, we only conducted experiments using a 13B-parameter white-box model, without exploring the impact of scaling to larger white-box models.

% \section*{References}

\medskip

% \today
{\small
 \bibliographystyle{ieee}
 \bibliography{main}
}
\newpage

\newpage
\appendix
\section{Appendix}
\subsection{Initialization Prompt}
% \newpage
 To guide ChatGPT in generating high-quality instructions, we utilized a structured initialization prompt. This prompt instructs the model to formulate 40 distinct instructions, each beginning with the phrase “The instruction was to,” while varying in style, structure, and complexity. The instructions must serve as high-level guiding principles without explicitly referencing input-output examples. Instead, they should be diverse in phrasing and sentence construction to ensure robustness and generalizability.

The following is the complete initialization prompt along with representative \texttt{\{init\_token\}} examples from the \textit{taxonomy\_animal} task:

\begin{tcolorbox}[colback=white!97!gray, colframe=black, width=\textwidth, arc=2mm, boxrule=0.8pt, sharp corners]
\textbf{You are an expert in crafting instructions that guide a chatbot to generate specific outputs from given inputs.}  

Your task is to create 40 distinct instructions, each beginning with “The instruction was to” and varying in style, structure, and length. These instructions should not reference example inputs or outputs but should act as high-level guiding principles that, when paired with the right input, lead to the desired output. Ensure that each instruction differs significantly in wording, complexity, and sentence structure. Avoid mentioning inputs, outputs, or using explicit examples within the instructions.

Below are example input-output pairs used as \texttt{\{init\_token\}}:

\begin{verbatim}
Input: pink, shark, giraffe, truck, duck, monkey
Output: giraffe, monkey, shark, duck

Input: mango, camel, chicken, sock, frog, donkey
Output: donkey, chicken, camel, frog

Input: snail, chicken, potato, waiter, penguin, white
Output: chicken, snail, penguin

Input: snail, horse, green, blue, butterfly, scarf
Output: butterfly, horse, snail

Input: goat, pajamas, scarf, hippo, egg, whale
Output: hippo, whale, goat

Input: elephant, jacket, designer, jellyfish, gray, bee
Output: bee, jellyfish, elephant

Input: tortoise, camel, pink, fly, elephant, hat
Output: tortoise, elephant, camel, fly

Input: gray, swan, camel, businesswoman, whale, cook
Output: swan, whale, camel

Input: bread, sheep, brown, octopus, spider, spaceship, snail
Output: sheep, snail, octopus, spider

Input: parrot, sheep, goat, lion, orange, motorway
Output: goat, lion, sheep, parrot
\end{verbatim}
\end{tcolorbox}

\subsection{Task Difficulty Calculation}

To evaluate task difficulty in our benchmark, we compute difficulty scores based on multiple performance metrics across baselines. The final scores are shown in Table~\ref{table:task_difficulty}. Task difficulty is measured using three metrics: \textbf{Mean Error Difficulty}, which reflects the error rate across baselines where lower performance indicates higher difficulty; \textbf{Standard Deviation Difficulty}, which captures variability in performance, with higher variance suggesting greater difficulty; and \textbf{Combined Difficulty}, a weighted combination of the above metrics that balances error and variability:
\setlength{\abovedisplayskip}{1pt}  % 调整公式上方的间距
\setlength{\belowdisplayskip}{1pt}  % 调整公式下方的间距

\begin{equation}
    D_{\text{combined}}^{(i)} = \alpha \cdot ( 1 - \mu^{(i)} ) + \beta \cdot \sigma^{(i)},
\end{equation}
where
\begin{itemize}
    \item \( D_{\text{combined}}^{(i)} \): Combined difficulty score for the \( i \)-th sample/task.
    \item \( \mu^{(i)} \): Mean score of the \( i \)-th sample across evaluations (higher mean implies lower error).
    \item \( \sigma^{(i)} \): Standard deviation of scores for the \( i \)-th sample (quantifies performance variability).
    \item \( \alpha, \beta \): Weighting coefficients balancing error magnitude and variability (e.g., \(\alpha = 0.7\), \(\beta = 0.3\) in implementation).
\end{itemize}

\begin{table}[ht!]
\centering
\caption{Computed task difficulty scores using different metrics.}
\begin{tabular}{lccc}
\toprule
Task & Mean Error Difficulty & Standard Deviation Difficulty & Combined Difficulty \\
\midrule
active\_to\_passive & 0.0022 & 0.0031 & 0.0025 \\
antonyms & 0.2021 & 0.0674 & 0.1593 \\
auto\_categorization & 0.7632 & 0.0291 & 0.5552 \\
auto\_debugging & 0.7114 & 0.0465 & 0.5238 \\
cause\_and\_effect & 0.4188 & 0.0842 & 0.3678 \\
common\_concept & 0.8988 & 0.0209 & 0.6384 \\
diff & 0.6306 & 0.1941 & 0.5425 \\
first\_word\_letter & 0.0000 & 0.0000 & 0.0000 \\
informal\_to\_formal & 0.6025 & 0.0442 & 0.4579 \\
larger\_animal & 0.8692 & 0.0476 & 0.6438 \\
letters\_list & 0.0842 & 0.2062 & 0.1038 \\
negation & 0.2320 & 0.0155 & 0.1748 \\
num\_to\_verbal & 0.0006 & 0.0011 & 0.0008 \\
object\_counting & 0.6178 & 0.0940 & 0.5084 \\
odd\_one\_out & 0.4124 & 0.1297 & 0.3647 \\
orthography\_starts\_with & 0.4660 & 0.0658 & 0.3919 \\
periodic\_elements & 0.8906 & 0.0332 & 0.6647 \\
rhymes & 0.5418 & 0.3612 & 0.4624 \\
second\_word\_letter & 0.4586 & 0.3405 & 0.4023 \\
sentence\_similarity & 0.9148 & 0.1422 & 0.6932 \\
sentiment & 0.8924 & 0.0152 & 0.6399 \\
singular\_to\_plural & 0.1960 & 0.2227 & 0.2053 \\
sum & 0.3320 & 0.1654 & 0.3127 \\
synonyms & 0.7555 & 0.0973 & 0.5782 \\
taxonomy\_animal & 0.4553 & 0.1652 & 0.4043 \\
translation\_en-de & 0.7362 & 0.0821 & 0.5825 \\
translation\_en-es & 0.6874 & 0.0598 & 0.5404 \\
translation\_en-fr & 0.7178 & 0.0690 & 0.5644 \\
word\_sorting & 0.5380 & 0.1172 & 0.4414 \\
word\_unscrambling & 0.4923 & 0.1073 & 0.4152 \\
\bottomrule
\end{tabular}
\label{table:task_difficulty}
\end{table}

\newpage
\subsection{Baseline Performance Analysis Across Tasks and Difficulty}

We present a bubble chart visualization comparing task performance across APE, EvoPrompt, INSTINCT, InstructZero, and our proposed method. Each bubble represents a task, with the x-axis denoting task IDs, y-axis showing normalized scores (0–1), and size indicating task difficulty.
\begin{figure*}[t!]

    \centering
    
    \includegraphics[width=0.4\textwidth]{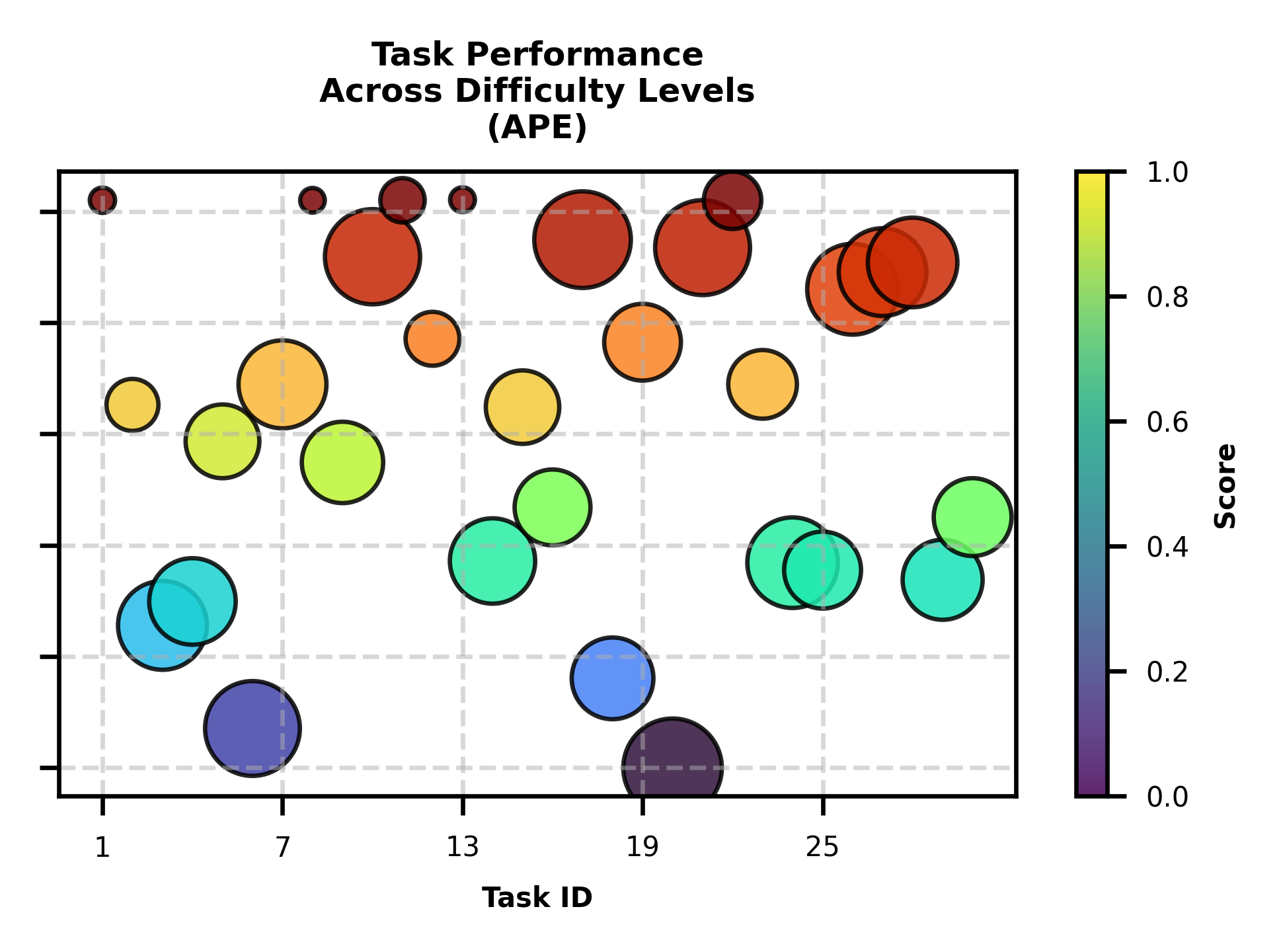}
    \includegraphics[width=0.4\textwidth]{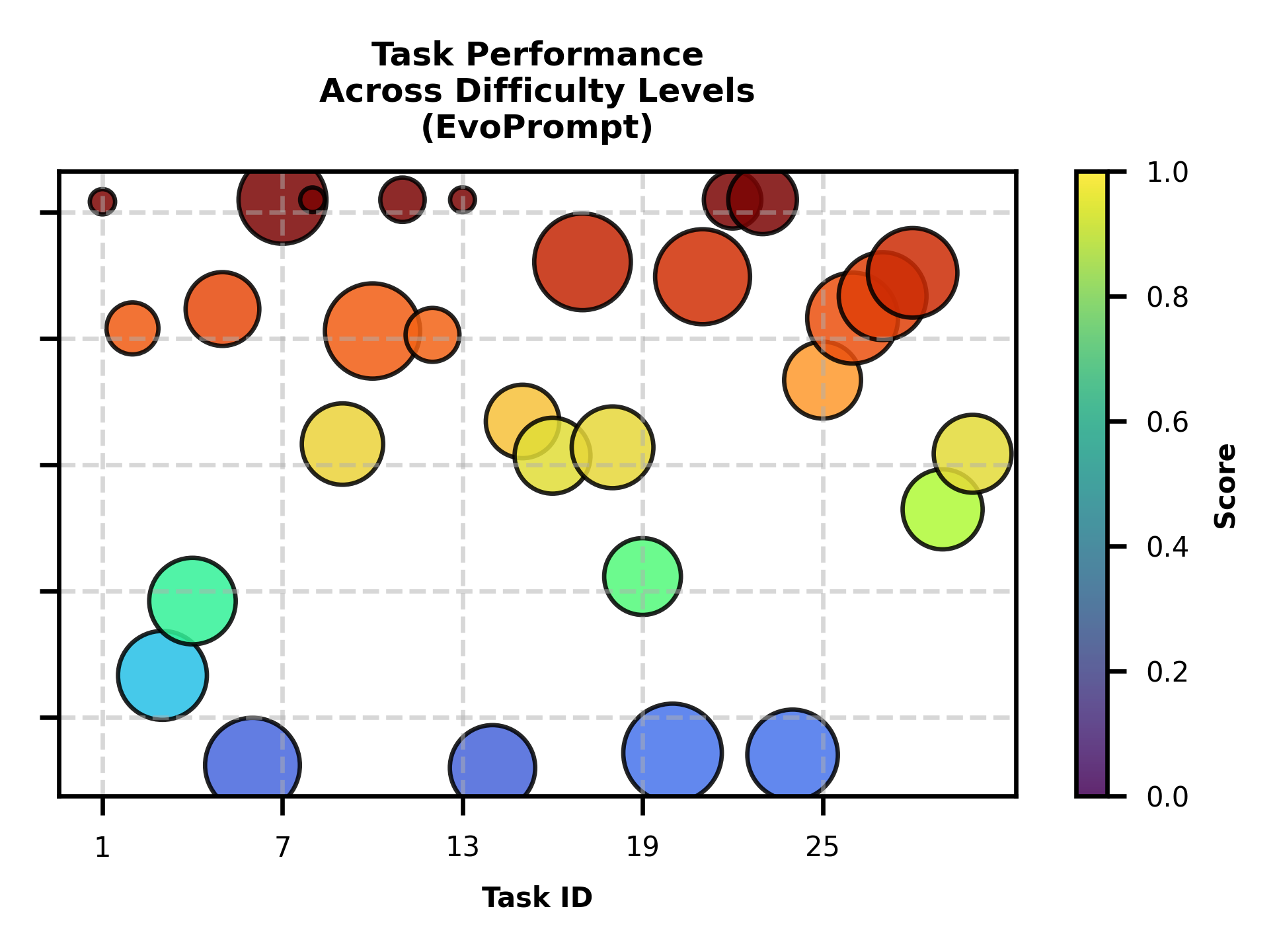} \\
    \includegraphics[width=0.4\textwidth]{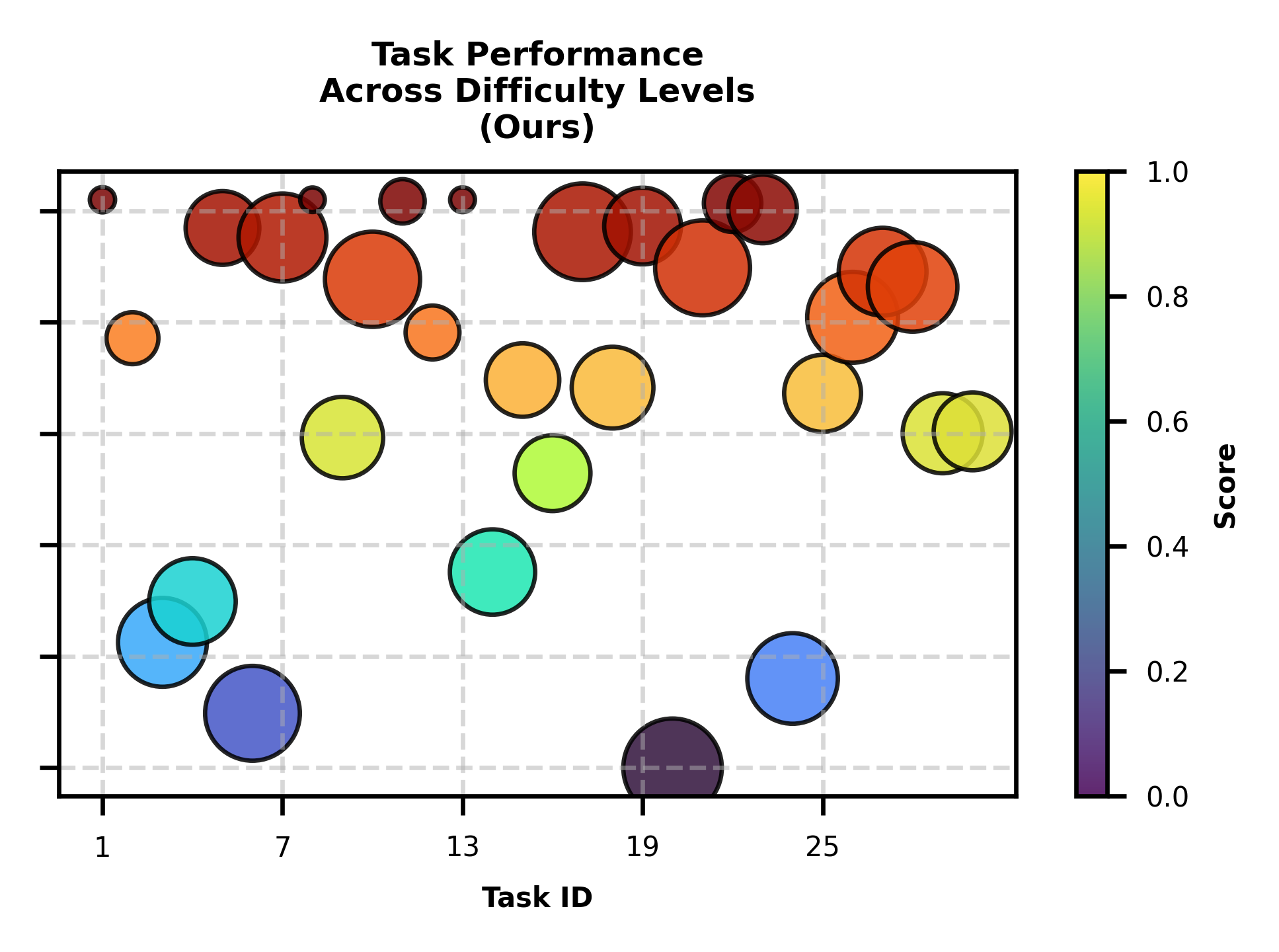} 
    \includegraphics[width=0.4\textwidth]{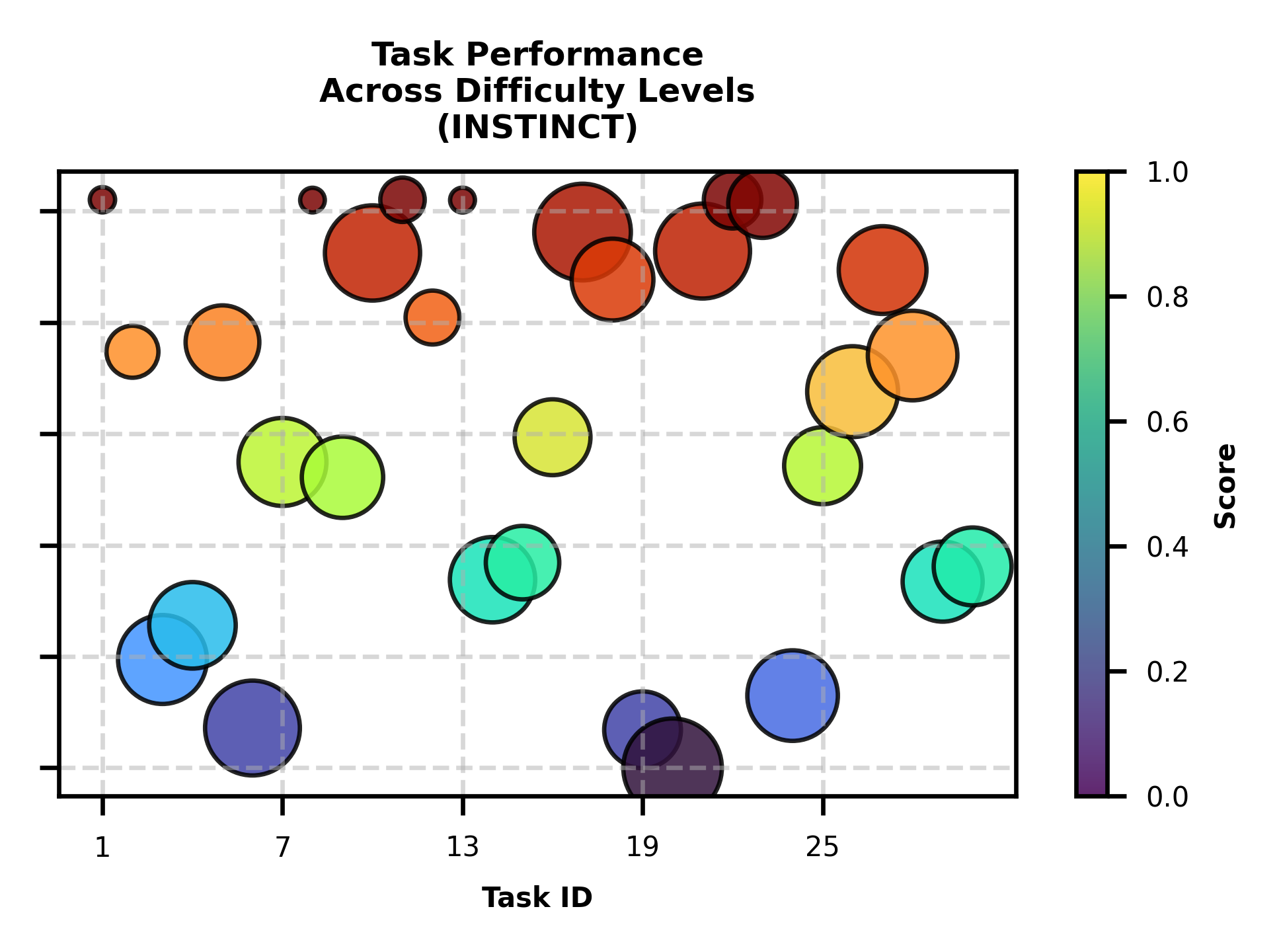}
    \includegraphics[width=0.4\textwidth]{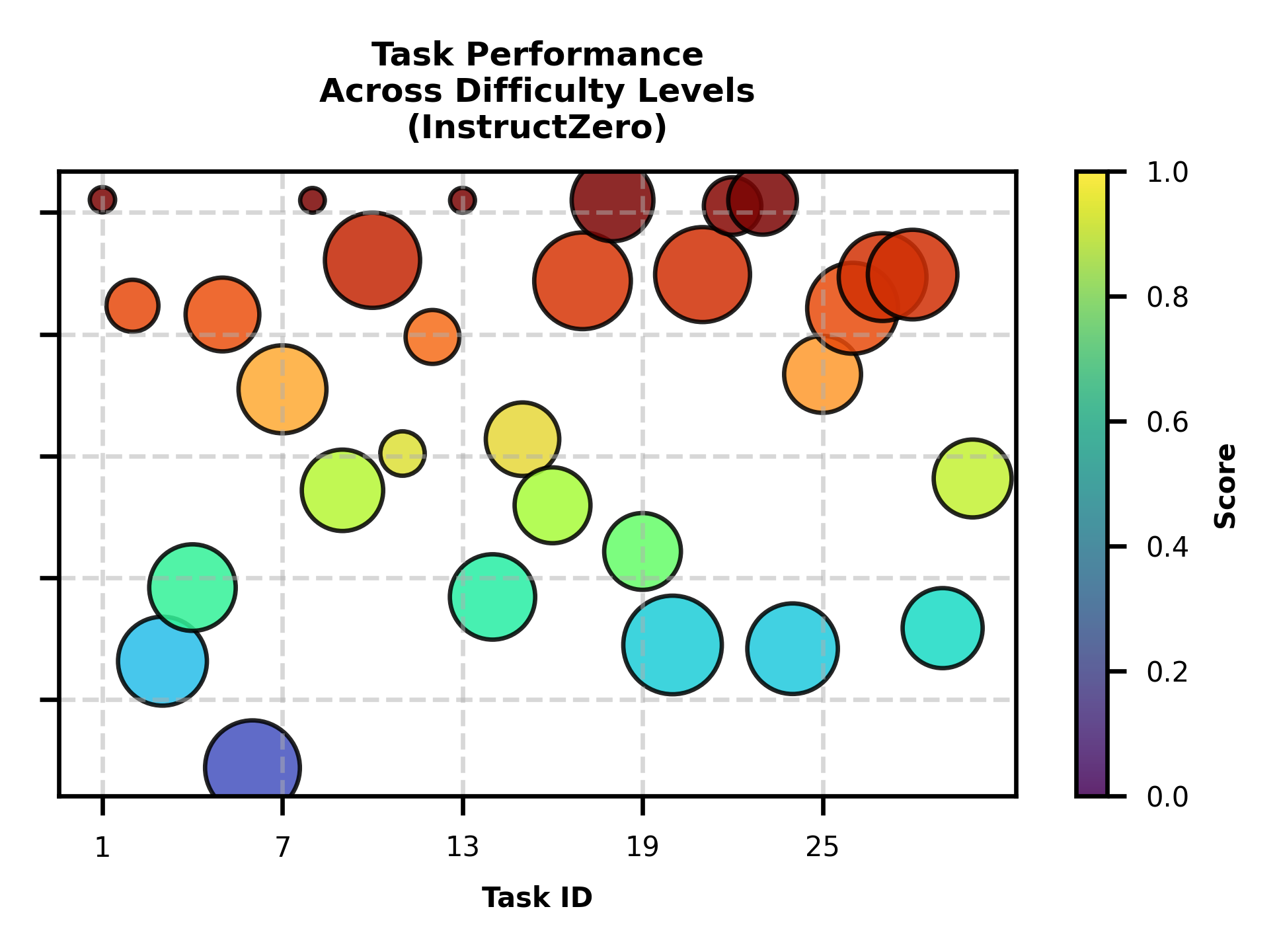}

    \caption{Bubble chart comparison of task performance across different baselines: APE, EvoPrompt, INSTINCT, and InstructZero. The chart shows performance across tasks with varying difficulty levels, where bubble size represents task difficulty and color indicates performance scores.  }
    \label{fig:4bubbles}
\end{figure*}
Distinct performance patterns emerge across baselines. APE exhibits high variance, excelling in some tasks while underperforming in others. EvoPrompt maintains stable performance, particularly in moderate-difficulty tasks. INSTINCT performs consistently but struggles with complex tasks requiring strong latent representations. InstructZero benefits from adaptive refinement, though challenges persist in lower-scoring regions. Our method demonstrates improved performance across a wider range of tasks, achieving higher scores in more challenging cases, as evident in Figure~\ref{fig:4bubbles}.

Across methods, higher task difficulty generally corresponds to greater performance fluctuations, highlighting the need for robust instruction optimization strategies balancing semantic adaptability and task-specific refinements. Our approach leverages both black-box initialization and white-box refinement, enabling more consistent optimization across diverse tasks.

\newpage

\newpage
\subsection{Best Instruction Across 30 Tasks}

\begin{table}[ht!]
    \centering
    \caption{Best-performing instructions for different tasks. Each instruction represents the most effective formulation found during iterative optimization.}
    \label{tab:best_instructions}
    \renewcommand{\arraystretch}{1.1}
    \begin{tabular}{lp{10.2cm}}
        \toprule
        \textbf{Task} & \textbf{Best Instruction} \\
        \midrule
        active\_to\_passive & Rewrite the sentence to shift attention from the initiator of the action to the entity upon which the action is performed. \\
        antonyms & Identify the logical inverse of the provided term and articulate it as a word. \\
        auto\_categorization & Identify the common category or theme shared by the listed items. \\
        auto\_debugging & Use the Python interpreter to execute each line of code and print the output. \\
        cause\_and\_effect & Highlight the statement that provides insight into the initial conditions affecting the plot. \\
        common\_concept & Highlight a physical process that both items undergo. \\
        diff & Calculate the difference by executing a basic subtraction operation. \\
        first\_word\_letter & Output the first letter of each inputted word. \\
        informal\_to\_formal & Restructure the sentence to align with formal language standards. \\
        larger\_animal & Determine which of the given options is more substantial in size and pick it. \\
        letters\_list & Provide a spaced sequence of characters that reconstructs the original word faithfully. \\
        negation & Reject the factual claim in the sentence through the use of negation. \\
        num\_to\_verbal & Convert numbers in the range 1-100000 to their word form. \\
        object\_counting & Count the number of items in the input and output the count. \\
        odd\_one\_out & Identify the singular element among a collection that fits a specific conceptual or thematic criterion. \\
        orthography\_starts\_with & Locate the term within the sentence that is associated with the single-letter hint. \\
        periodic\_elements & Determine the element’s name based on its location in the periodic chart. \\
        rhymes & Create a text that is as close to the original as possible, but with the given output. \\
        second\_word\_letter & Extract the second letter from the sequence and display it. \\
        sentence\_similarity & Use the output of the previous model as a guide to determine the correctness of the generated text. \\
        sentiment & Give each input-output pair a positive or negative evaluation based on whether the output was an improvement or a worsening compared to the input. \\
        singular\_to\_plural & Create a program that takes a word as input and outputs its plural form. \\
        sum & Add the first number to the second number. \\
        synonyms & Create a program that takes input from the user and outputs a word that is similar in meaning to the input word. \\
        taxonomy\_animal & Extract and sequentially arrange animal-related terms, omitting any non-animal words. \\
        translation\_en-de & Translate the given input and output words into German, and create a new rule for each input-output pair. \\
        translation\_en-es & Translate the words from English to Spanish. \\
        translation\_en-fr & Identify the French equivalent for the provided English term. \\
        word\_sorting & Place the words in order from the first letter of the alphabet to the last. \\
        word\_unscrambling & Transform the scrambled input into a recognizable word. \\
        \bottomrule
    \end{tabular}
\end{table}

\newpage
\subsection{More Details on Ablation Experiment}
\begin{table}[ht!]
    \centering
    \caption{Performance comparison across different tasks with varying weight settings and neural network depth. The last row reports the mean performance.}
    \label{tab:moredetails app}
    \renewcommand{\arraystretch}{1.1}
    \begin{tabular}{lccc|cc}
        \toprule
        \multirow{2}{*}{\textbf{Task}} & \multicolumn{3}{c|}{\textbf{SimReg-weight}} & \multicolumn{2}{c}{\textbf{NN-depth}} \\
        \cmidrule(lr){2-4} \cmidrule(lr){5-6}
        & \textbf{weight = 0.01} & \textbf{weight = \(10^{-10}\)} & \textbf{Ours} & \textbf{depth = 1} & \textbf{depth = 5} \\
        \midrule
        active\_to\_passive & 1.0000 & 1.0000 & 1.0000 & 0.9867 & 0.9967 \\
        antonyms & 0.7633 & 0.7467 & 0.7533 & 0.7700 & 0.8033 \\
        auto\_categorization & 0.3333 & 0.3167 & 0.2200 & 0.1400 & 0.2300 \\
        auto\_debugging & 0.3333 & 0.3333 & 0.2917 & 0.2917 & 0.2083 \\
        cause\_and\_effect & 0.9067 & 0.8667 & 0.9467 & 0.9467 & 0.9067 \\
        common\_concept & 0.0868 & 0.0930 & 0.0956 & 0.0817 & 0.0727 \\
        diff & 0.9667 & 0.9700 & 0.9300 & 0.9667 & 0.9433 \\
        first\_word\_letter & 0.9967 & 0.9767 & 1.0000 & 0.9933 & 0.9867 \\
        informal\_to\_formal & 0.5144 & 0.5804 & 0.5791 & 0.4142 & 0.5024 \\
        larger\_animal & 0.7500 & 0.8900 & 0.8567 & 0.7733 & 0.8333 \\
        letters\_list & 1.0000 & 0.9967 & 0.9933 & 1.0000 & 1.0000 \\
        negation & 0.7833 & 0.8300 & 0.7633 & 0.8267 & 0.7767 \\
        num\_to\_verbal & 0.9933 & 0.9933 & 1.0000 & 0.9933 & 1.0000 \\
        object\_counting & 0.2733 & 0.3567 & 0.3433 & 0.2867 & 0.3000 \\
        odd\_one\_out & 0.6133 & 0.6400 & 0.6800 & 0.6133 & 0.5800 \\
        orthography\_starts\_with & 0.6500 & 0.5733 & 0.5167 & 0.6200 & 0.4900 \\
        periodic\_elements & 0.8800 & 0.9467 & 0.9400 & 0.9267 & 0.8133 \\
        rhymes & 0.4067 & 0.4567 & 0.6667 & 0.4167 & 0.6333 \\
        second\_word\_letter & 0.9167 & 0.7033 & 0.9500 & 0.7833 & 0.8400 \\
        sentence\_similarity & 0.0000 & 0.0000 & 0.0000 & 0.0000 & 0.0000 \\
        sentiment & 0.8867 & 0.6467 & 0.8767 & 0.9000 & 0.8600 \\
        singular\_to\_plural & 0.9967 & 0.9767 & 0.9900 & 0.9700 & 0.9900 \\
        sum & 0.9733 & 0.9667 & 0.9800 & 0.8967 & 0.9367 \\
        synonyms & 0.1767 & 0.1867 & 0.1567 & 0.1133 & 0.1100 \\
        taxonomy\_animal & 0.4567 & 0.7433 & 0.6567 & 0.4233 & 0.4200 \\
        translation\_en-de & 0.8033 & 0.8167 & 0.7900 & 0.7967 & 0.7733 \\
        translation\_en-es & 0.8500 & 0.8667 & 0.8700 & 0.8567 & 0.8567 \\
        translation\_en-fr & 0.8433 & 0.8033 & 0.8433 & 0.8033 & 0.8133 \\
        word\_sorting & 0.5967 & 0.6067 & 0.5867 & 0.6233 & 0.5733 \\
        word\_unscrambling & 0.4733 & 0.4900 & 0.5900 & 0.5767 & 0.4667 \\
        \rowcolor{gray!20} \textbf{MEAN} & 0.6742 & 0.6791 & 0.6955 & 0.6597 & 0.6572 \\
        \bottomrule
    \end{tabular}
\end{table}
Table~\ref{tab:moredetails app}presents the performance comparison across different tasks with varying weight settings for SimReg and neural network depth. The results show the effect of different configurations on task performance, highlighting the performance variation as the SimReg-weight and neural network depth are adjusted.

For SimReg-weight, with a value of 0.01, the method achieves strong performance across a variety of tasks, with notable scores in tasks like \textit{active\_to\_passive} (1.0000), \textit{first\_word\_letter} (1.0000), and \textit{letters\_list} (1.0000). However, when the SimReg-weight is set to a very small value (10$^{-10}$), we observe a noticeable drop in performance for tasks such as \textit{auto\_categorization} (0.2200) and \textit{cause\_and\_effect} (0.0956), indicating that a low SimReg-weight adversely affects performance in tasks that require fine-tuned semantic alignment.

In terms of neural network depth, we observe that increasing the depth from 1 to 5 generally leads to an improvement in performance across several tasks. For instance, \textit{periodic\_elements} increases from 0.9267 to 0.9400, and \textit{word\_unscrambling} improves from 0.5900 to 0.6233 when the network depth increases. This suggests that a deeper network helps capture more complex patterns and improves performance, particularly in tasks requiring higher-level reasoning and structural alignment.

The last row in Table 5 shows the mean performance across all tasks. Our method, with a SimReg-weight of 0.01 and depth of 5, achieves a mean score of 0.6955, which is the highest among the configurations, demonstrating the optimal combination of hyperparameters for task performance. In contrast, the lowest mean performance (0.6572) occurs when the SimReg-weight is set to 10$^{-10}$ and depth is set to 1, indicating that both lower weight values and shallow network depths hinder the model's ability to generalize effectively across tasks.

These results underline the importance of selecting appropriate hyperparameters for optimal task performance, and our findings suggest that a higher SimReg-weight and deeper network depth lead to more robust performance across a variety of tasks.

\begin{figure}[t!]
    \centering
    \includegraphics[width=0.5\textwidth]{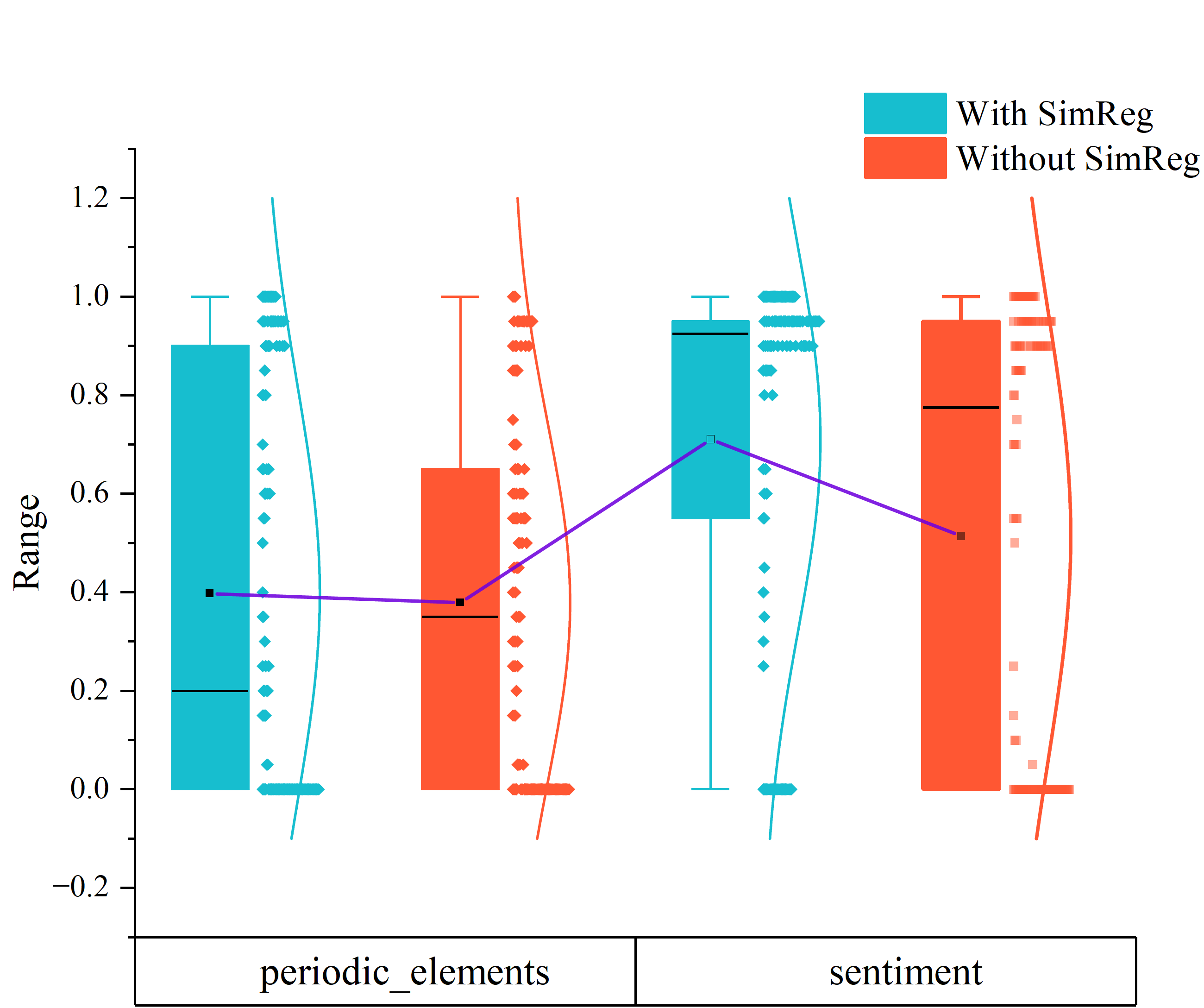} 
    \caption{Box plot comparing instruction optimization performance with and without similarity regularization (SimReg).}
    \label{fig:abaltion-loos2}
\end{figure}

\begin{figure*}[t!]

    \centering

    \includegraphics[width=0.32\textwidth]{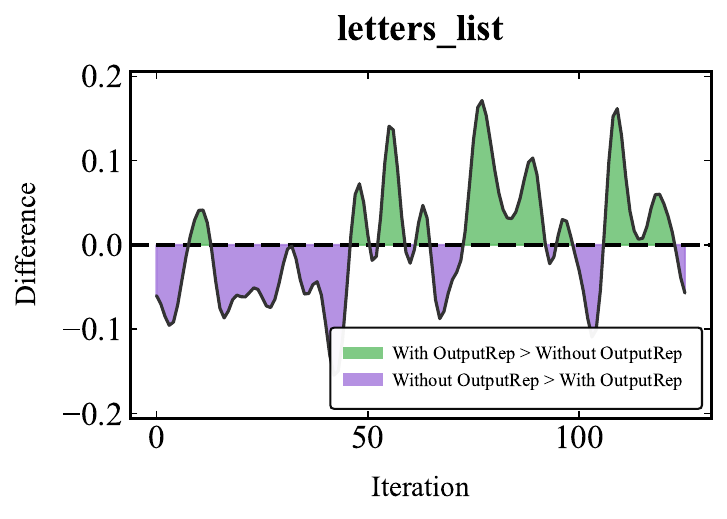}
    \includegraphics[width=0.32\textwidth]{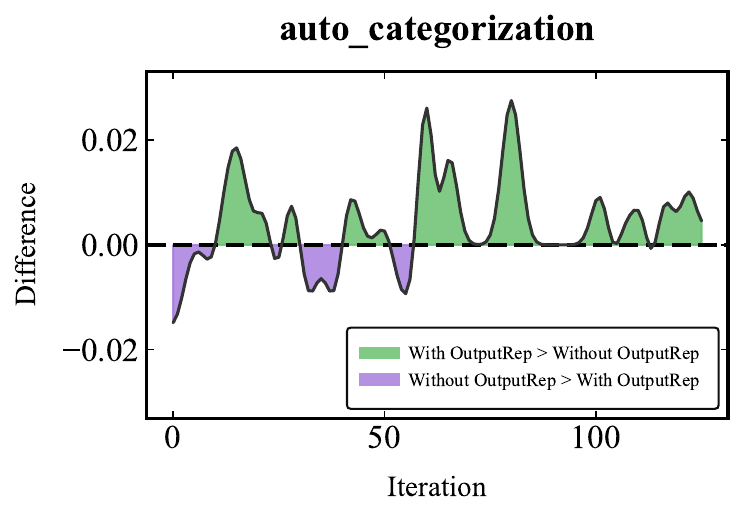}
    \includegraphics[width=0.32\textwidth]{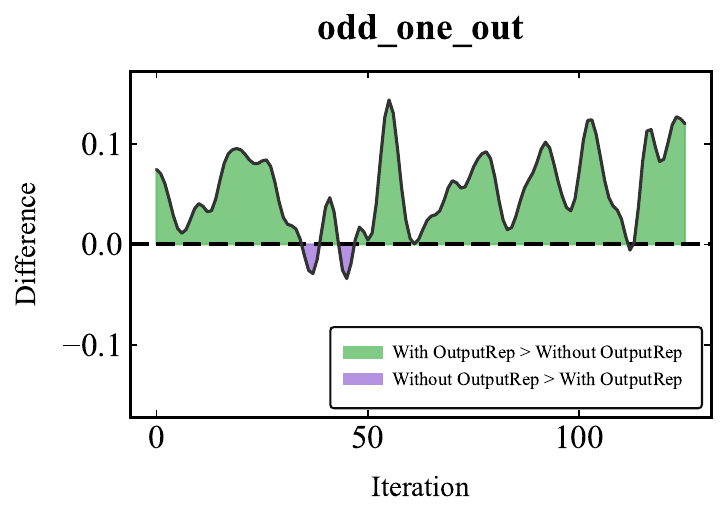} 
\caption{Ablation study on the impact of output representation. }
\label{fig:ablation--outrep}
\end{figure*}
%%%%%%%%%%%%%%%%%%%%%%%%%%%%%%%%%%%%%%%%%%%%%%%%%%%%%%%%%%%%
% \input{NList}
\end{document}